\documentclass[11pt]{article}

\usepackage[margin=1in]{geometry}
\usepackage[utf8]{inputenc}
\usepackage[T1]{fontenc}
\usepackage{lmodern}
\usepackage{microtype}
\usepackage{amsmath,amssymb,amsfonts}
\usepackage{graphicx}
\usepackage{booktabs}
\usepackage{multirow}
\usepackage{array}
\usepackage{longtable}
\usepackage{caption}
\usepackage{xcolor}
\usepackage{enumitem}
\usepackage[section]{placeins} % keep floats within their section (prevents pdfLaTeX cascading all figures to the end)
\usepackage{tikz}
\usetikzlibrary{arrows.meta,positioning,fit,backgrounds,calc,shapes.geometric}
\usepackage{listings}
\usepackage[numbers,sort&compress]{natbib}
\usepackage[hidelinks]{hyperref}
\usepackage{cleveref}
\usepackage[normalem]{ulem} % [normalem] prevents the package from changing italics to underlines globally

\definecolor{accent}{HTML}{2F5FD6}
\definecolor{good}{HTML}{1A9F6E}
\definecolor{bad}{HTML}{D23B48}
\definecolor{warn}{HTML}{B8860B}
\definecolor{ink}{HTML}{1C2030}
\definecolor{boxbg}{HTML}{F3F5FB}
\definecolor{boxln}{HTML}{D5DAEA}

\lstdefinestyle{qep}{
  basicstyle=\ttfamily\footnotesize,
  breaklines=true,
  columns=fullflexible,
  keepspaces=true,
  showstringspaces=false,
  frame=single,
  rulecolor=\color{boxln},
  backgroundcolor=\color{boxbg},
  framesep=5pt,
  xleftmargin=4pt,
  xrightmargin=2pt,
}
\newcommand{\code}[1]{\texttt{\small #1}}

\newsavebox{\defectcardbox}
\newenvironment{defectcard}[1]{%
  \par\noindent
  \setlength{\fboxsep}{6pt}%
  \setlength{\fboxrule}{0.6pt}%
  \begin{lrbox}{\defectcardbox}%
    \begin{minipage}{\dimexpr\linewidth-2\fboxsep-2\fboxrule\relax}%
    \colorbox{bad}{\color{white}\bfseries\footnotesize\strut #1}%
    \par\smallskip\footnotesize
}{%
    \end{minipage}%
  \end{lrbox}%
  \fcolorbox{boxln}{boxbg}{\usebox{\defectcardbox}}%
  \par
}
\newcommand{\defectfield}[2]{{\fontsize{8.5}{9.8}\selectfont{\normalfont\bfseries #1:}\ \itshape #2\par}\vspace{2pt}}

\title{\textbf{Reasoning Jury: Multi-Model Consensus for Evaluating Reasoning Traces}}

\author{%
    \shortstack[c]{%
      Congchao Wang, Diwakar Singh, Qiaozi Gao, Spyros Matsoukas, Yang Liu, \\[4pt]
      Mahdi Namazifar\thanks{Corresponding author:
        \href{mailto:mahdinam@amazon.com}
        {\texttt{mahdinam@amazon.com}}}\\[9pt]
      \textbf{Amazon AGI}%
    }%
  }

\date{}

\begin{document}
\maketitle

\begin{abstract}
Improving reasoning LLMs requires the ability to judge the quality of long reasoning traces for effective reasoning data curation, strong training signals during reinforcement learning, and an in-depth understanding of reasoning behaviors during model performance evaluation. 
Additionally, surfacing reasoning mistakes that the model makes would enable improving the model's performance at runtime through providing feedback.
Due to the difficulty of this complex task on long reasoning traces, single-model judges (even frontier models) do not do well at identifying reasoning defects. Additionally, leveraging frontier models during online training of reasoning LLMs is generally prohibited due to guardrails in terms of use.
In this work, we introduce \emph{Reasoning Jury}, a system that replaces the single judge with a jury of LLMs and a moderated consensus mechanism, to improve the fidelity of judgments for identifying reasoning defects.
% In reasoning jury, defects of a reasoning trace and their severity are surfaced through a deliberation where a moderator conducts a discussion amongst the jury where the jurors critique each other's judgments and get to modify their initial votes.
The moderator derives a consensus through deliberation amongst jurors or consolidation of judgements. We show that Reasoning Jury with a jury of open-weight models is able to significantly outperform frontier models (Opus, Sonnet, and Gemini) at correctly identifying reasoning defects. Besides accuracy performance improvements, the aggregated cost of the jury (initial verdicts, deliberations, consolidation, etc.) is a fraction ($8$ to $16\%$) of the cost of running frontier models in LLM-as-a-judge setup. We also show how these judgements can be leveraged to understand failure modes of reasoning LLMs on benchmarks, which allows much deeper understanding of a model's performance. 
\end{abstract}

% ============================ INTRODUCTION ============================
\section{Introduction}
\label{sec:intro}

Reliable evaluation of reasoning traces  supports several stages of development of reasoning LLMs.
 Supervised fine-tuning
depends on high-quality reasoning traces and filtering out low quality data; reinforcement learning depends on reward signals that separate good reasoning from bad; and performance evaluation of reasoning LLMs and error analysis
depend on diagnosing reasoning outputs. All of these depend on the ability to decide how good a reasoning trace actually is. Unlike non-reasoning models where evaluation  concerns short and self-contained answers, in reasoning LLMs the content being judged has shifted to long chain-of-thought traces that include tens to
hundreds of interdependent steps, in which a single bad step could propagate through
everything that follows, and in which spotting the defect at all is often highly challenging. In this setting the evaluation is no longer only ``is the final
answer right'' but ``which reasoning step went wrong, and how badly''. 
% This impacts numerous dimensions of the quality of a reasoning LLM, including its token efficiency, proper self-checks, and calibration of confidence.
Such fine-grained evaluations can provide actionable signals for improving other dimensions of reasoning quality, including token efficiency, self-checking behavior, and confidence calibration.

\begin{figure}[t]
\centering
\includegraphics[width=\linewidth]{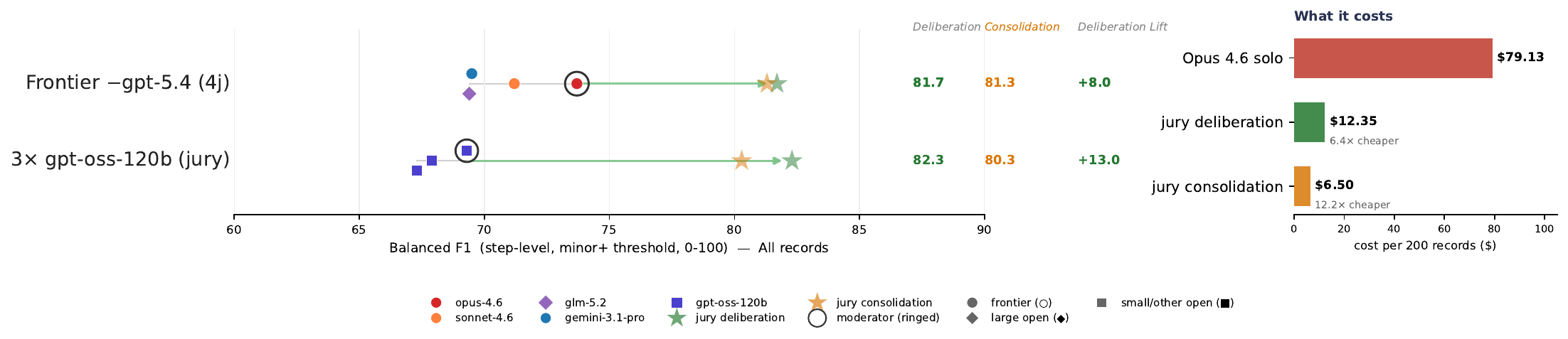}
\caption{Comparing Balanced F1 scores of reasoning defect detection on Hard2Verify benchmark with frontier models (opus-4.6, sonnet-4.6, gemini-3.1-pro) jury vs an open-weight jury (3 instances of gpt-oss-120b). Dots are individual jurors and stars are jury results (green for deliberation and orange for consolidation). The frontier models outperform gpt-oss-120b individually. However Reasoning Jury of gpt-oss-120b outperforms these frontier models by up to 12 points. The dollar cost of this Reasoning Jury in deliberation mode is 15.6\% and in consolidation mode is 8\% of the cost of opus-4.6 at this task (see \Cref{tab:results-cost}).}
\label{fig:h2v_frontier_vs_oss}
\end{figure}

Identifying defective reasoning traces, especially those with higher severity enables effective training data filtering for mid-training and SFT stages. Additionally identifying where exactly such defects occur enables mechanisms to rewrite the defective parts of a reasoning trace either leveraging expert human annotators or using LLMs. 
Moreover, such high-fidelity defect detection could be leveraged during RL to steer reasoning behaviors of the models. Approaches such as NuRL \cite{chen2026nudging} (uses offline-generated hints during generation), scaf-GRPO \cite{zhang2026scafgrposcaffoldedgrouprelative} (uses in-prompt hints when training stagnation is detected), SDPO \cite{hubotter2026reinforcement} (uses natural language feedback as hints), etc. are RL approaches that rely on rich signals beyond an outcome score. These signals could include a binary defect based score, natural language hints to avoid defects, or specific tokens where a defect occurred; and RL could strongly benefit from high-fidelity and rich reasoning defect detection. 
Besides model training, new inference time scaling approaches such as \cite{agrawal2026optimize} leverage evaluation of the model's outputs in an iterative loop as input back to the model to improve the model's performance at run time. Following this approach, identified reasoning defects could be fed back to the model to correct its reasoning mistakes and improve its performance.
Also in model evaluations, instead of only reporting accuracy scores on benchmarks, reasoning defects, their distributions, and major failure modes could also be reported that would guide the training process and model improvements.

The dominant solution for such judgements is LLM-as-a-judge with a single strong model \citep{zheng2023judging,gu2024survey}. This is adequate on short outputs, but breaks down precisely on long reasoning due to the long context of reasoning traces and the complexity of reasoning defect detection task (results in \Cref{sec:results:single}). 
% For instance frontier models disagree with \emph{themselves} when asked to re-judge the same long reasoning trace, so their verdict is not even stable under resampling (\textcolor{red}{ref to be added}). 
% Additionally, on objective-correctness benchmarks such as JudgeBench \citep{tan2025judgebench} (which test reasoning, math, and coding) frontier judges score only slightly above random. 
For such challenging tasks, a single model brings a single perspective to the judging of a reasoning trace, and whatever it misses or misjudges stays incorrect. Here a natural remedy mirrors how human institutions handle high-stakes judgements, namely convene a \emph{jury}, where several independent judges drawn from diverse model families have decorrelated blind spots, so a misjudgement by one is often caught by another. 
This is a growing line of work where panels of diverse LLMs (and debate among them) improve evaluation over any single judge
\citep{verga2024replacing,du2024improving,chen2025majeval}. We build on this line and specialize it to the problem of detecting step-grounded defects in long reasoning traces.

To this end, we introduce \emph{Reasoning Jury}. The main assumption of the approach 
is that every reasoning trace that is input to Reasoning Jury is broken into reasoning ``steps'' demarcated by 
\code{[STEP-x]} where \code{x} is a step counter. In turn, every verdict highlighted by Reasoning Jury is grounded
to a specific \code{[STEP-x]} marker in the trace, which makes each judgement
auditable (a human can check it) and actionable (it points at the exact step where a defect occurs). To identify reasoning defects, Reasoning Jury runs a 
two-phase process over a panel, where in Phase 1 independent judgements are obtained from the jurors and in Phase 2 a consensus is reached based on these verdicts.

In Phase~1, each juror reads the problem and the step-segmented reasoning trace and, \emph{independently}, emits a structured list of defects. Each defect comes with 1) a
self-contained 
% \code{what\_went\_wrong} 
description of the defect, 2)
% \code{statement\_refs}  which is 
a list of \code{[STEP-x]} in which the defect is, 3) severity
% \code{impact} which indicates severity (\code{neutral}/\code{minor}/\code{major}/\code{fatal}) 
of the defect, and 4) detailed 
% \code{evidence} 
evidence of the defect in the reasoning trace.
A representative defect example for a reasoning trace for solving a math Olympiad problem  is shown in \Cref{fig:intro-example}.
As a result, the jurors commit decorrelated views before any of them can anchor on the others. 

In Phase~2, a consensus from these verdicts is derived. This phase has two modes, namely Consolidation and Deliberation. In Consolidation mode the original content along with Phase 1 verdicts are passed to a judge, and the judge is asked to consolidate them. In Deliberation mode, a moderator runs a multi-turn debate in which jurors argue over these defects until a consensus is reached or the deliberation is stalled. The moderator decides which juror speaks on each turn, what the juror should address, and when the discussion has converged enough to terminate. The moderator then synthesizes the panel's consensus from the full transcript of deliberations.
It is worth note that the moderator is not given the original problem, reasoning trace, or candidate solution. It nevertheless sees content-rich juror arguments, maintains a claim-level consensus state, selects speakers, and determines termination. Thus, the design separates direct access to the source material from control of the discussion. We treat trace withholding as a design heuristic intended to limit direct re-adjudication, not as a structural guarantee of unbiased moderation. Its causal effect is not isolated in our experiments.
By design, Phase~1 and Phase~2 preserve a compatible core output schema for individual defect findings, while Phase~2 adds consensus and deliberation metadata where applicable.

\begin{figure}[t]
\centering
\begin{defectcard}{Fatal defect}
\defectfield{\uline{Problem}}{Let $k\geq 2$ be an integer. Determine all sequences of positive integers $a_1,a_2,\ldots$ for which there exists a monic polynomial $P$ of degree $k$ with non-negative integer coefficients such that $P(a_n)=a_{n+1}a_{n+2}\cdots a_{n+k}$ for every integer $n\geq 1$.}
\defectfield{\uline{What went wrong}}{   In \code{[STEP-3]} the trace claims that, for the characteristic polynomial $r^k+r^{k-1}+\cdots+r-k=0$, ``all roots other than $r=1$ must have absolute value strictly less than $1$.'' This is false: for $k{=}2$ the polynomial factors as $(r-1)(r+2)$, giving a root $r=-2$ with $|r|=2>1$. The trace's triangle-inequality argument only rules out roots on the unit circle other than $1$; it never excludes $|r|>1$. This false claim is exactly what the derivation needs to conclude that $d_n=a_{n+1}-a_n$ is eventually constant, so the rest of the case rests on a false foundation.}
\defectfield{\uline{Statement references}}{   \code{["STEP-3"]}}
\defectfield{\uline{Impact}}{\textcolor{bad}{   Fatal}}
\defectfield{\uline{Evidence}}{   ``\,So all roots other than $r=1$ must have absolute value strictly less than $1$. The general solution to the recurrence is $d_n = A\cdot 1^n+\sum_i P_i(n)r_i^n$ where $|r_i|<1$. As $n\to\infty$, $d_n\to A$.\,''}
\end{defectcard}
\caption{A representative defect emitted by a juror in Phase~1, on an olympiad
problem from Hard2Verify. The juror grounds its judgement to a specific
\code{[STEP-x]}, gives a self-contained explanation of the mistake, tags a severity,
and quotes the trace as evidence.}
\label{fig:intro-example}
\end{figure}

To evaluate Reasoning Jury we use reasoning defect benchmarks Hard2Verify \citep{lin2025hard2verify} and DeltaBench \cite{he-etal-2025-large}.  On these benchmarks we show that a jury of open-weight models could easily outperform frontier models. Additionally, although jury consolidation or deliberation consumes significantly more tokens than a single model judge,  we show that Reasoning Jury costs a fraction compared to running a frontier LLM as a judge.  As there are legal and practical limitations on using closed-weight frontier models for training LLMs, the jury of open-weight models with high fidelity would provide a flexible path to use such juries for LLM training.

Our benchmark evaluations measure whether the jury localizes defective steps, but
they do not directly validate the factual accuracy of
\code{what\_went\_wrong}, the calibration of severity labels, or the general
downstream usefulness of these fields. A comprehensive assessment of those
properties across models and tasks remains future work. We nevertheless
provide an initial task-based test of whether richer defect descriptions help
a model repair its reasoning. For each AIME2026 problem, we generate $64$
solutions with nemotron-3-super ($1{,}920$ traces), evaluate every trace with
the jury, and ask the same model to retry each of the $877$ traces flagged with
at least one non-neutral defect. Supplying the \code{what\_went\_wrong}
diagnosis and severity raises retry accuracy to $76.2\%$ from $71.2\%$ with step locations
alone. \Cref{sec:profiling:retry} summarizes the
experiment, and \Cref{app:retry} provides the full design and results.

% \paragraph{Contributions} 
% In this work we focus on identifying defects in reasoning traces grounded in reasoning steps with details of each defect including what went wrong and severity of the defect. This task enables enhanced reasoning training data filtering, provides strong learning signals in RL, and enables in-depth analysis of reasoning behaviors of the model at evaluation time. 
% For this task we introduce Reasoning Jury, which is a multi-model consensus framework that replaces single-model LLM-as-a-judge with a jury of models, to increase fidelity of this complex judgement task. We show that leveraging open-weight models in Reasoning Jury can outperform frontier models at both performance and cost by a wide margin. For instance, such a jury could outperform a single frontier LLM judge (e.g., opus-4.6, sonnet-4.6, or Gemini-3.1-Pro) at this task (\Cref{fig:h2v_frontier_vs_oss}) at 8\% of the dollar cost (\Cref{tab:results-cost}). 
% Using an iterative algorithm we also leverage defects surfaced by Reasoning Jury to create a taxonomy of reasoning defects as a diagnostic instrument. We then use this taxonomy to profile reasoning defects of a model at evaluation time on a benchmark to highlight reasoning mistakes and failure modes of the model. This model benchmark evaluation approach takes the evaluation beyond single accuracy numbers towards detailed analyses and insights on the types of mistakes that the model makes and provides actionable directions on how to improve the model's performance. 

% ============================ METHOD ============================
\section{Reasoning Jury}
\label{sec:method}

\subsection{Segmenting the trace into steps}
\label{sec:method:steps}

Grounding of detected defects of a reasoning trace would require some way to reference 
where exactly the defect occurs within a long reasoning trace. Splitting a long reasoning 
trace into \emph{reasoning steps} would enable this grounding where a juror can cite for example \code{[STEP-14]} when claiming a defect. In order to achieve this reasoning step segmentation the vast majority of work 
in the literature insert a step boundary at every double newline 
(the pattern \code{\textbackslash n\textbackslash n})~\citep{lightman2023lets,zheng2024processbench,zou2025reasonfluxprm,cheng2025pure,sharma2026prism}. Although simple, this
approach has its drawbacks, including potentially placing step boundaries in 
the middle of a code or pseudo-code snippet, a multi-line algebraic derivation, or a 
simple enumeration of different cases. Additionally if a model simply uses a single 
newline instead of double, or in general uses line breaks less frequently this approach becomes less robust.
Another approach for this could be using a strong LLM as a one-shot segmenter, 
where an LLM is prompted to add step markers, and produce
semantically coherent steps for a long reasoning trace. But the same long-content 
self-instability that motivates this paper applies here as well where segmentation 
is not stable across re-samples, and most importantly the large LLM may not remain fully
faithful to the main reasoning text and makes modifications to it, and as a result, the 
segmented reasoning trace is different from the original reasoning trace.
Ideally a dedicated, specialized reasoning step segmentation model would address this 
need; but absent that, we use the pattern of an end-of-sentence character followed by 
\code{\textbackslash n\textbackslash n} (regex pattern \code{(?<=[.!?])\textbackslash s*\textbackslash n\textbackslash s*\textbackslash n+}) 
to segment a reasoning trace to reasoning steps. The addition of end-of-sentence 
characters improves the robustness of the pattern for traces including code and multi-step algebraic derivations.

\subsection{Reasoning Jury Pipeline}
\label{sec:method:pipeline}

\subsubsection{Phase 1: independent judgement}
All jurors receive the same prompt that includes the original problem, the step-annotated trace, and
the final solution. Each independently returns a JSON object of
verdicts, where each verdict carries 
\code{what\_went\_wrong}, a self-contained description of the defect;
\code{statement\_refs}, a list of steps that the defect refers to; \code{impact}, the severity of the defect ($\mathtt{neutral}\!\mid\!\mathtt{minor}\!\mid\!\mathtt{major}\!\mid\!\mathtt{fatal}$); and
\code{evidence}, the supporting evidence of the identified defect. The full prompt can be found in \Cref{app:phase1prompt}. It enforces a \emph{genericness test} (``could
this comment apply to a different problem with no edits? if so, rewrite it'')
and a \emph{specificity self-check} (1--5; include only issues scoring
$\geq 4$) to suppress vague, non-grounded criticism. 
\code{what\_went\_wrong} tries to capture weaknesses in the reasoning trace by pointing to bad reasoning moves. It is intentionally kept high-level and open-ended with requirements on specificity to not limit the jurors in identifying different kinds of defects. 
Phase 1 calls fan out in
parallel, and the pipeline proceeds as long as a configurable minimum number of
jurors succeed.

\subsubsection{Phase 2: Consensus}
Consensus from Phase 1 independent judgements is reached in two different modes, namely Consolidation and Deliberation. 

\paragraph{Consolidation.}
In this mode a judge is asked to perform the task as Phase 1, except it is also given all the verdicts from Phase 1. The judge (moderator) is given instructions on how to verify, merge, and fill in the gaps in the Phase 1 judgements. The final output of Consolidation is in the format of Phase 1 outputs. The full prompt is provided in \Cref{app:consolidationprompt}.

\paragraph{Deliberation.} The jury deliberations start based on Phase 1 verdicts.
The moderator (see full prompt in \Cref{app:moderatorprompt}) is intentionally designed to be blind to the problem, the
reasoning trace, and the candidate solution. It sees only the deliberation transcript, and its
role is purely procedural (manage deliberation turns, surface disagreements, detect convergence),
never speculating about reasoning trace content or hinting at the right answer. Deliberation is
organized into logical rounds, within each of which every juror must speak at least
once. Each turn proceeds in three parts. 

First, the moderator selects the next speaker
and issues a process-oriented instruction, prioritising jurors involved in unresolved
disagreements, recalling any juror silent for two or more turns, and asking the chosen
juror to clarify a specific point and defend or concede it.
Second, the selected juror (see full prompt in \Cref{app:phase2contribution}) re-reads 
the problem and the reasoning trace, as well as the moderator's instructions, and contributes a
natural-language argument (agreeing, disagreeing with evidence, raising a new defect, or
conceding) addressing the moderator's asks. 
Third, the moderator folds the contribution into a running consensus state and checks for
deliberation termination conditions. 
The loop terminates on a full logical round with no new
substantive argument, on universal agreement, or on detected cycling, with a hard cap on total turns as a backstop regardless. 
A final extraction call to the moderator receives the Phase 1 judgements, the full transcript, and the moderator's running deliberations. The moderator then synthesizes the
\emph{consensus defects} (a defect reaches consensus if a dynamically
computed majority endorsed it, or if it was raised and never contested), a
\emph{confidence} in $[0,1]$ reflecting the degree of agreement,
\emph{dissenting views} that did not reach consensus, and the
\emph{termination reason}. The majority threshold is computed from the number
of jurors that actually succeeded, so partial failures do not silently change
the voting rule.

The integrity of the deliberation rests on a deliberate division of labor
among non-juror roles. 
In deliberation mode, if the moderator could see the trace, it would inevitably form opinions about
the defects, and those opinions would leak into \emph{whom it calls on} and
\emph{what it tells them to address}, making it a covert extra juror with
the unique power to suppress dissent by never giving it the floor. By
restricting the moderator to the \emph{conversation only} (speaker names and
message content, never the problem or trace), its decisions are necessarily
about \emph{argument dynamics} (who has not spoken, which disagreement is
unresolved, whether the round produced anything new) not about content.

% ============================ EXPERIMENTAL SETUP ============================
\section{Experimental Setup}\label{sec:setup}

We now describe how we evaluate Reasoning Jury, the benchmarks and the evaluation procedures.

\subsection{Benchmarking Accuracy of Reasoning Jury}\label{sec:setup:benchmark}

Since reasoning defect detection is an under-studied task in the published literature, there are very few available datasets to leverage for benchmarking this task. Our evaluations mainly use Hard2Verify \citep{lin2025hard2verify} which is a human-annotated, step-level verification benchmark for open-ended advanced mathematics competitions with 200 records. These records have approximately 780 gold error steps (on average around 9.3 steps per record, of which around 3.9 are labeled erroneous). Additionally we also use DeltaBench \cite{he-etal-2025-large}, which is a benchmark for this task covering STEM, coding, and general reasoning with 1,236 records, and includes defect localization over reasoning traces. The reasoning traces in these benchmarks are substantially longer than those of prior step-level benchmarks such as ProcessBench \citep{zheng2024processbench}. For that reason we do not consider using ProcessBench in this work (discussed further in \Cref{app:h2vdifficulty}). For cost related concerns, the vast majority of our evaluations are done on the smaller benchmark Hard2Verify, but we also provide detailed results for a subset of evaluations for DeltaBench.

In order to evaluate reasoning jury on these benchmarks we take the rich and detailed step-level detected defects and turn them into step level binary signals. If a step was highlighted in a detected defect (in the \code{statement\_refs} field of the defect) with severity \emph{minor}, \emph{major}, or \emph{fatal}, that step is labeled as 1; otherwise step labels are 0. Following Hard2Verify paper we mainly look at Balanced F1, the harmonic mean of error-recall and specificity, micro-averaged over steps. Alongside Balanced F1 we report Balanced Accuracy, Accuracy, Precision, Recall, and F1. Note that we evaluate only step-level defect localization. This scoring discards what went wrong, evidence, confidence, dissent, and the distinction among minor, major, and fatal severity. Consequently, these experiments do not evaluate explanation factuality, evidential support, severity calibration, semantic deduplication, or downstream usefulness.

The Hard2Verify paper uses a simple prompt to output a list of verdicts for each step (correct or incorrect) as well as a list of reasoning for each verdict. The prompt that we leverage in this paper is much more elaborate and produces verdicts with severity tags (which helps with reasoning data curation and filtering and provides reliable and detailed signals for RL), as well as mechanisms to ensure defect specificity that, intuitively speaking, helps in root causing issues in other domains. The Hard2Verify paper reports Balanced F1 score of 85.8 with their prompt using gpt-5. We replicate that experiment with gpt-5.4 (which is the version we use in this paper) and we get the Balanced F1 score of 85.3. Using our prompt (\Cref{app:phase1prompt}) with gpt-5.4 we get the Balanced F1 score of 83.9. Based on these numbers and the additional utilities that our prompt provides, for the rest of the experiments we leverage our prompt for reasoning defect detection.

We score against the benchmarks' own gold step boundaries and gold error labels. We do \emph{not} re-segment solutions for scoring purposes. This keeps the step-level comparison apples-to-apples against the human annotation. 
As an additional point, the Hard2Verify authors evaluate 29 generative critics and process reward models, and the strongest process reward models score in the range 0.35--0.60 Balanced F1 at the step level. We cite this range as context for what a strong model achieves on this task, without importing any per-model figure.

\subsection{Evaluation Configuration}
In all of our experiments we set the reasoning effort of all LLM calls to ``High''. Temperature 1.0 is used across all LLM calls. For gpt-5.4 calls (except gpt-oss-120b) we use Amazon Mantle API, for opus-4.6 and sonnet-4.6 we use Amazon Bedrock API, for gemini-3.1-pro we use OpenRouter API, and for all open-weight models we serve them on a local cluster using vLLM.

% ============================ RESULTS ============================
\section{Results}\label{sec:results}
Most of our results in this section are on the Hard2Verify benchmark, and we also report results on DeltaBench for some of the key experiments in \Cref{sec:results:deltabench}. Throughout this section we mostly report on Balanced F1 score at step level, and we provide full tables in \Cref{app:fulltables}. 
For Hard2Verify, because each score is measured on roughly 200 records, every number carries sampling uncertainty; we report 95\% bootstrap confidence intervals (10,000 record-level resamples) on the Balanced-F1.

\subsection{Single-model as a judge}\label{sec:results:single}

We first establish how well individual models detect reasoning defects. From \Cref{fig:h2v-jury}, it is clear that gpt-5.4 at $83.9$ Balanced F1 is a saturated outlier that sits roughly ten points clear of the next best single model, opus-4.6 at $73.7$. The remainder of the frontier tier (sonnet-4.6 at $71.2$, gemini-3.1-pro at $69.5$) and the strongest open model glm-5.2 at $69.4$ performs lower than opus-4.6. The open-weight models as a whole span a wide band, from roughly $50$ to $70$.

The contrast between the two leading models is notable. gpt-5.4's lead is through a balance between  precision and recall in the high $70$s to low $80$s (\Cref{tab:results-jury-solo}), whereas opus-4.6 has a high precision ($81.7$) but a low recall ($62.5$), and it misses many genuine defects. This asymmetry foreshadows the aggregation results below, where combining jurors mostly recovers recall.

\subsection{Jury versus a single judge}\label{sec:results:jury}

\begin{figure}[t]
\centering
\includegraphics[width=\linewidth]{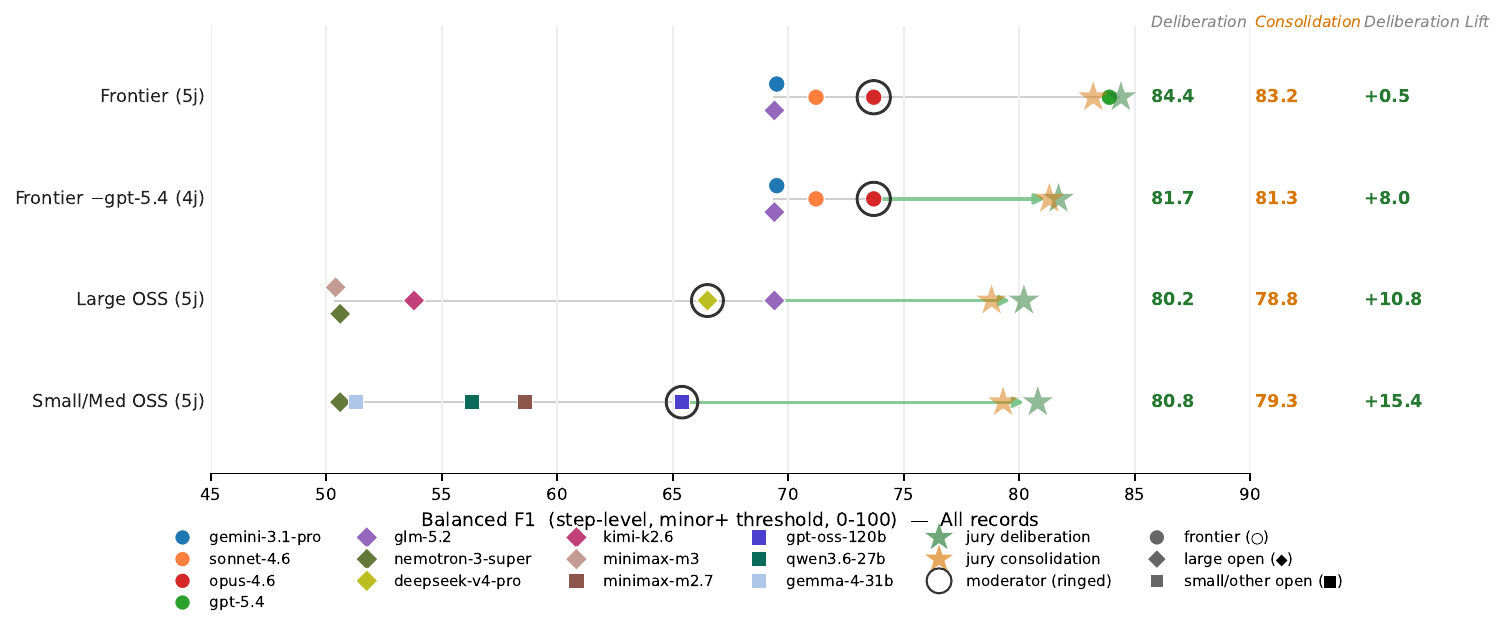}
\caption{Solo jurors (dots) versus deliberated consensus (star) for each jury, with the lift over best juror on the right. The open-source juries gain the most; the full Frontier jury is near-saturated by gpt-5.4, so its lift is small, whereas removing gpt-5.4 restores a large lift. Exact numbers, with $95\%$ confidence intervals, in \Cref{tab:results-jury-solo}.}
\label{fig:h2v-jury}
\end{figure}

Next we move on from a single judge to a jury with a consensus verdict. 
For these experiments we use Hard2Verify benchmark and we create the following juries. (1) Frontier which includes frontier proprietary models gpt-5.4~\citep{openai2026gpt54,openai2026gpt54docs}, opus-4.6~\citep{anthropic2026opus46}, sonnet-4.6~\citep{anthropic2026sonnet46}, gemini-3.1-pro~\citep{googledeepmind2026gemini31pro}. We also include glm-5.2~\citep{glm5team2026glm5,zai2026glm52fp8} which is the largest open-weight model in our mix of models to the Frontier jury. (2) Frontier$-$gpt-5.4 which is the Frontier jury excluding gpt-5.4. Through this jury we study the removal of the outsized performance of gpt-5.4 from the jury. (3) Large OSS, which includes the largest open-weight models, namely glm-5.2, deepseek-v4-pro~\citep{deepseekai2026deepseekv4pro}, kimi-k2.6~\citep{moonshotai2026kimik26}, nemotron-3-super~\citep{nvidia2025nemotron3}, and minimax-m3~\citep{minimaxai2026minimaxm3}. (4) Small/Medium OSS models, which include qwen3.6-27b~\citep{qwenteam2026qwen36}, nemotron-3-super, gpt-oss-120b~\citep{openai2025gptoss}, gemma-4-31b~\citep{gemmateam2026gemma4}, minimax-m2.7~\citep{minimaxai2026minimaxm27}.
\Cref{tab:results-jury} reports the consolidation and deliberated final verdict for the four juries across 6 metrics, and \Cref{fig:h2v-jury} places each jury's consensus against its constituent jurors' Balanced F1 scores. 
From the figure it is clear that at $84.4$ Balanced F1 the Frontier jury is near-saturated by gpt-5.4's solo score of $83.9$, so the jury adds almost nothing on top of its dominant member. In Frontier$-$gpt-5.4 we remove the dominant gpt-5.4 from the jury while keeping the remaining four jurors' Phase 1 judgements fixed, and a Balanced F1 of $81.7$ is achieved, which is 8 points above its best juror opus-4.6, and it lands within 3 points of the gpt-5.4 anchored jury. Here a conclusion is that when one juror already saturates the task, consensus tracks that juror, and when that is not the case, aggregation across the merely-strong jurors does improve the performance.

Among open-weight models, the deliberating OSS juries substantially outperform every constituent juror. The strongest individual OSS jurors achieve only $65.4$--$69.4$ Balanced F1, whereas the Small/Medium and Large OSS juries reach $80.8$ and $80.2$, gains of $15.4$ and $10.8$ points, respectively. These gains are driven primarily by improved recall (\Cref{tab:results-jury-solo}), while maintaining precision near $80$. Notably, the Small/Medium OSS jury outperforms opus-4.6, sonnet-4.6, and gemini-3.1-pro by $7.1$--$11.3$ points and performs on par with the Large OSS jury.

Across all four juries, deliberation yields numerically higher Balanced F1 than consolidation, with differences ranging from $0.4$ to $1.5$ points (\Cref{tab:results-jury}). For the OSS juries, the distinction is primarily a precision--recall trade-off: deliberation raises recall from $71.3$ to $75.0$ for Large OSS and from $71.3$ to $77.4$ for Small/Medium OSS, whereas consolidation raises precision from $79.6$ to $81.1$ and from $78.3$ to $82.9$, respectively. Thus, deliberation recovers more true defects, while consolidation produces more conservative verdicts.

The Balanced-F1 confidence intervals in \Cref{tab:results-jury-solo} reflect sampling uncertainty over the approximately $200$ Hard2Verify records. Single-model estimates have
wider intervals ($\pm3$ to $\pm7$ points), while the deliberated consensus intervals range from $\pm2.5$ to $\pm3.1$ points. The exception is gpt-5.4, whose solo interval
($83.9\pm2.8$) is comparable to the Frontier consensus. Full per-juror intervals are provided in \Cref{tab:results-jury-solo} (\Cref{app:fulltables}).

\begin{table}[t]
\centering
\caption{Jury verdict for four juries under both Phase 2 modes: deliberation and consolidation. Both modes share identical Phase 1 judgements per jury; for Frontier $-$gpt-5.4 the Phase 1 judgements are inherited from the Frontier run (dropping only the gpt-5.4 seat), so that row isolates exactly the effect of removing the dominant juror. All six metrics are step-level, with severity minor and above. OSS juries lift Balanced F1 $10.8$ to $15.4$ over their best solo juror under deliberation. Consolidation retains most of that lift giving up $0.4$--$1.5$ Balanced F1 by trading recall for precision.}
\label{tab:results-jury}
\small
\setlength{\tabcolsep}{4.5pt}
\begin{tabular}{llcccccc}
\toprule
jury & Phase 2 & BalAcc & Bal-F1 & Acc & P & R & F1 \\
\midrule
\multirow{2}{*}{Frontier (5 jurors)} & Deliberation & $84.5$ & $84.4$\,{\tiny$\pm2.5$} & $84.1$ & $77.5$ & $87.1$ & $82.0$ \\
  & Consolidation & $83.2$ & $83.2$\,{\tiny$\pm2.7$} & $83.3$ & $78.7$ & $82.3$ & $80.5$ \\
\midrule
\multirow{2}{*}{Frontier $-$gpt-5.4 (4 jurors)} & Deliberation & $81.8$ & $81.7$\,{\tiny$\pm2.6$} & $82.3$ & $78.9$ & $78.2$ & $78.8$ \\
  & Consolidation & $81.4$ & $81.3$\,{\tiny$\pm2.7$} & $82.0$ & $78.8$ & $78.0$ & $78.4$ \\
\midrule
\multirow{2}{*}{Large OSS (5 jurors)} & Deliberation & $80.6$ & $80.2$\,{\tiny$\pm3.1$} & $81.5$ & $79.6$ & $75.0$ & $77.2$ \\
  & Consolidation & $79.7$ & $78.8$\,{\tiny$\pm3.1$} & $81.1$ & $81.1$ & $71.3$ & $75.9$ \\
\midrule
\multirow{2}{*}{Small/Medium OSS (5 jurors)} & Deliberation & $81.0$ & $80.8$\,{\tiny$\pm3.0$} & $81.6$ & $78.3$ & $77.4$ & $77.9$ \\
  & Consolidation & $80.3$ & $79.3$\,{\tiny$\pm3.2$} & $81.8$ & $82.9$ & $71.3$ & $76.6$ \\
\bottomrule
\end{tabular}
\end{table}

\subsection{How many jurors are needed?}\label{sec:results:knockout}

\begin{figure}[ht!]
\centering
\includegraphics[width=0.86\linewidth]{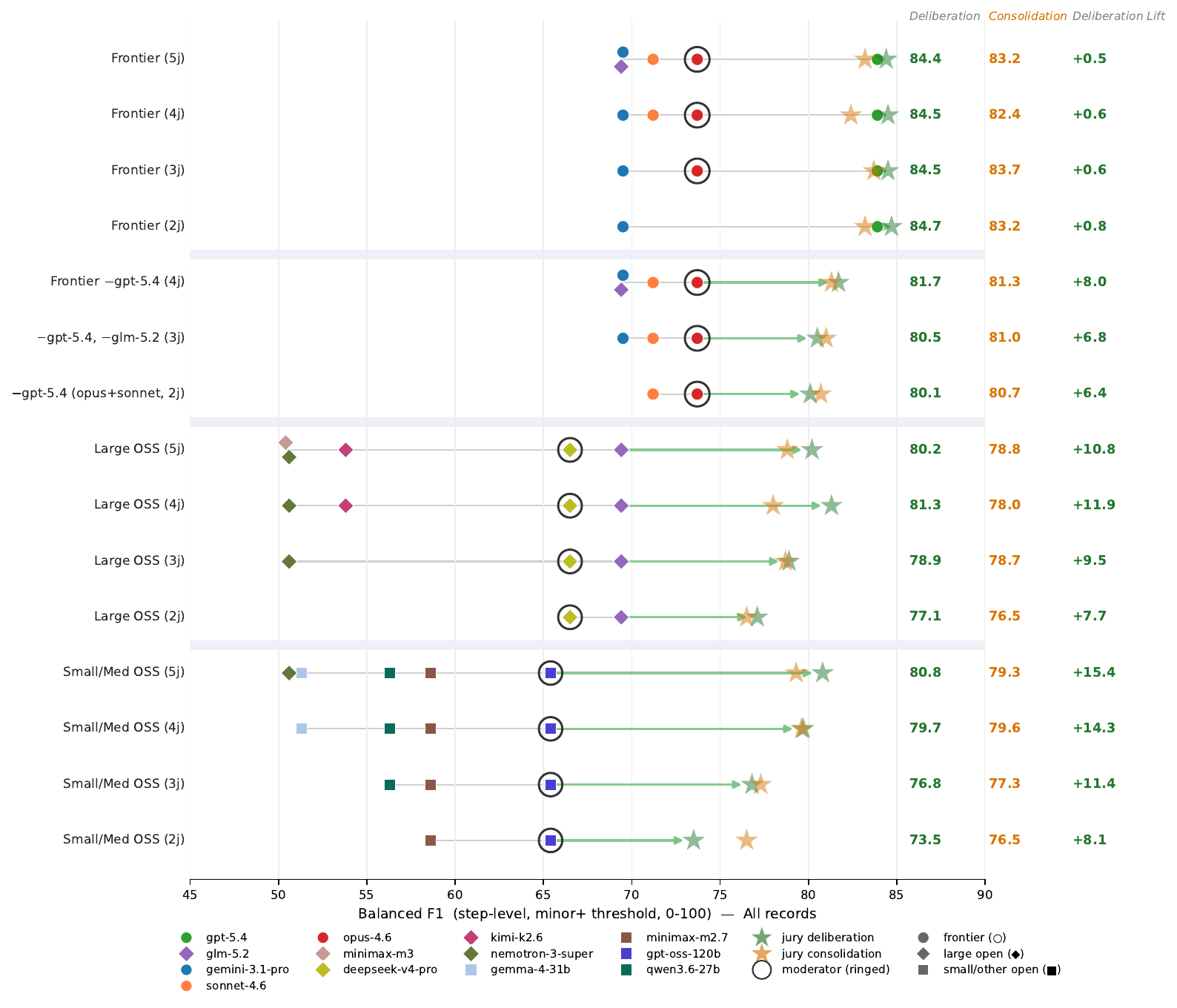}
\caption{Consensus versus solo jurors as each jury is reduced from five jurors down to two. Exact numbers in \Cref{tab:results-knockout}.}
\label{fig:h2v-knockout}
\end{figure}

To understand the impact of the size of the jury on its performance, for Hard2Verify we drop jurors from the jury one at a time. In each run, we drop the juror contributing the fewest \emph{unique} gold defective steps (steps that no other juror in the jury
also caught), computed from the preceding run's Phase 1 outputs. To isolate jury size from sampling noise, we hold Phase 1 fixed: each smaller jury reuses the exact independent Phase 1 judgements from the full jury's run, and only Phase 2 with both deliberation and consolidation are run over the surviving subset. \Cref{fig:h2v-knockout} depicts the results. The full Frontier jury performs the same from five to two jurors ($84.4 \to 84.5 \to 84.5 \to 84.7$ for deliberation) because gpt-5.4 carries the verdict regardless of what other models are in the jury. The Frontier $-$gpt-5.4 and Large OSS juries decline gently as jurors are removed, to $80.1$ and $77.1$ respectively with deliberation at two jurors. The Small/Medium OSS jury holds well through three jurors ($80.8 \to 79.7 \to 76.8$ with deliberation) and then drops more sharply to $73.5$ with two jurors.

% Regarding confidence of the Balanced-F1 column of \Cref{tab:results-jury-solo}, two things stand out. First, single-model detectors carry wide intervals ($\pm3$ to $\pm7$ points) that widen as accuracy drops, so a weak juror's score is not only lower but substantially less certain. Second, the deliberated consensus consistently yields the narrowest interval of any configuration in each panel ($\pm2.5$ to $\pm3.1$), a 30–45\% reduction relative to the average juror—evidence that deliberation improves not just the accuracy of the estimate but its stability. The lone exception is gpt-5.4, whose solo interval ($83.9\pm2.8$) is already as tight as the consensus and overlaps it almost entirely, which is exactly why the Frontier jury it anchors shows little room to improve. Full per-juror confidence intervals for all four panels are given in \Cref{tab:results-jury-solo} (\Cref{app:fulltables}).
  
Taken together, three jurors provide a reasonable operating point for the Frontier, Frontier$-$gpt-5.4, and Large OSS panels, remaining within $1.3$ points of the largest jury. For Small/Medium OSS, four jurors appear preferable: reducing from five to four costs only $1.1$ points, whereas reducing to three costs $4.0$ points.

\subsection{Which model should moderate?}\label{sec:results:moderator}

\begin{figure}[t]
\centering
\includegraphics[width=0.82\linewidth]{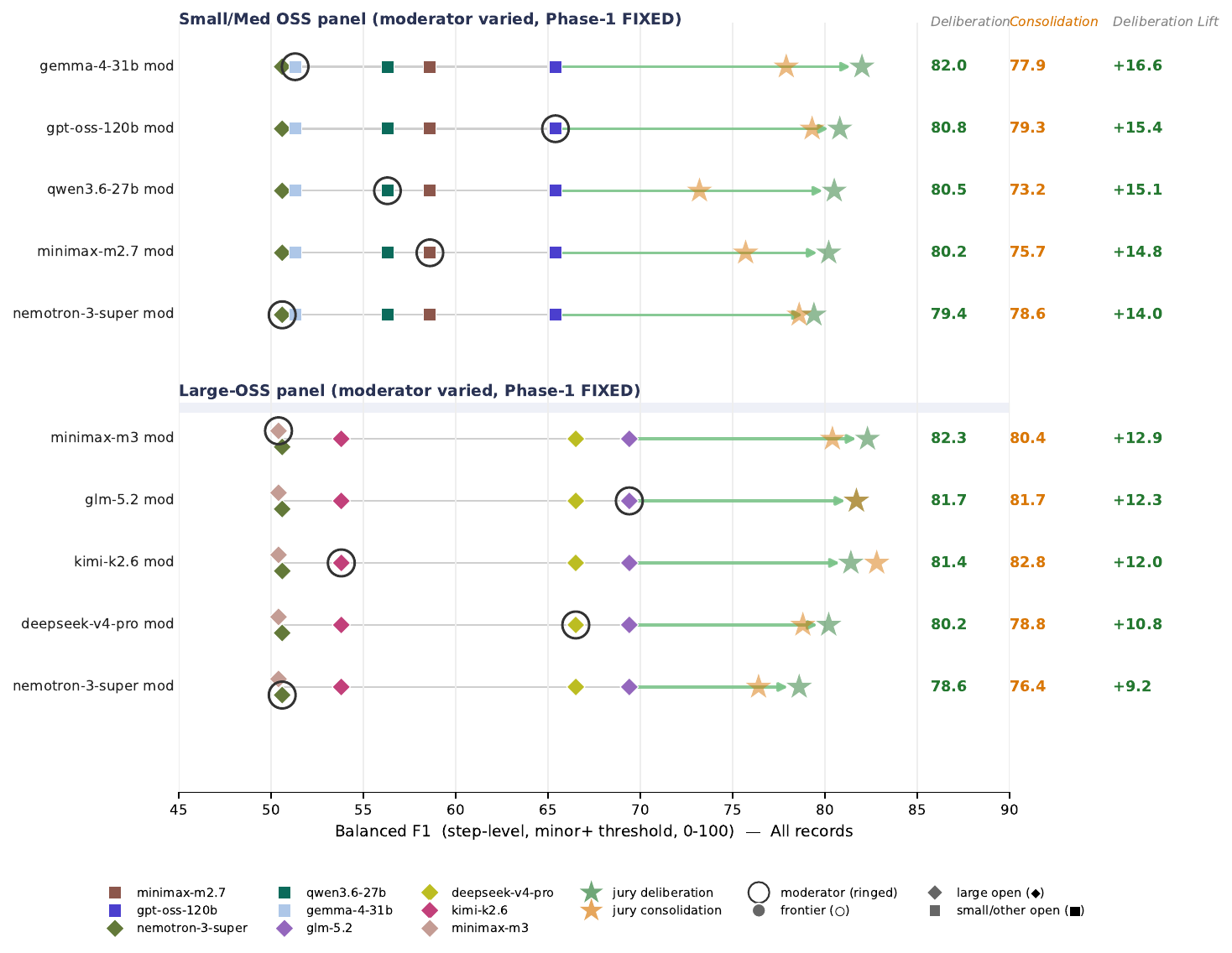}
\caption{Moderator choice changes deliberation by $2.6$--$3.7$ points and consolidation by $6.1$--$6.4$ points. Deliberation values are shown in \Cref{tab:results-moderator};  consolidation values are shown in the figure.}
\label{fig:h2v-modsweep}
\end{figure}

The moderator runs the jurors' arguments and issues the final verdict, so a question is how much the moderator choice matters when the jury itself is held fixed. \Cref{fig:h2v-modsweep} sweeps the moderator over each jury while leaving the jurors unchanged. Moderator choice with deliberation moves consensus by $3.7$ and $2.6$ points for the two OSS juries, namely Large OSS and Small/Medium OSS. With consolidation, the corresponding ranges are $6.4$ and $6.1$ points, respectively, which shows that consolidation is more sensitive to the choice of moderator. In consolidation mode we see that a weak moderator (e.g., qwen3.6-27b) could cause a $7$ point drop in the performance of the jury compared to deliberation.
The results show that the moderator plays a key role in the performance of the jury, and it should therefore be chosen carefully and with caution.

\subsection{Does jury diversity matter?}\label{sec:results:diversity}

\begin{figure}[t]
\centering
\includegraphics[width=\linewidth]{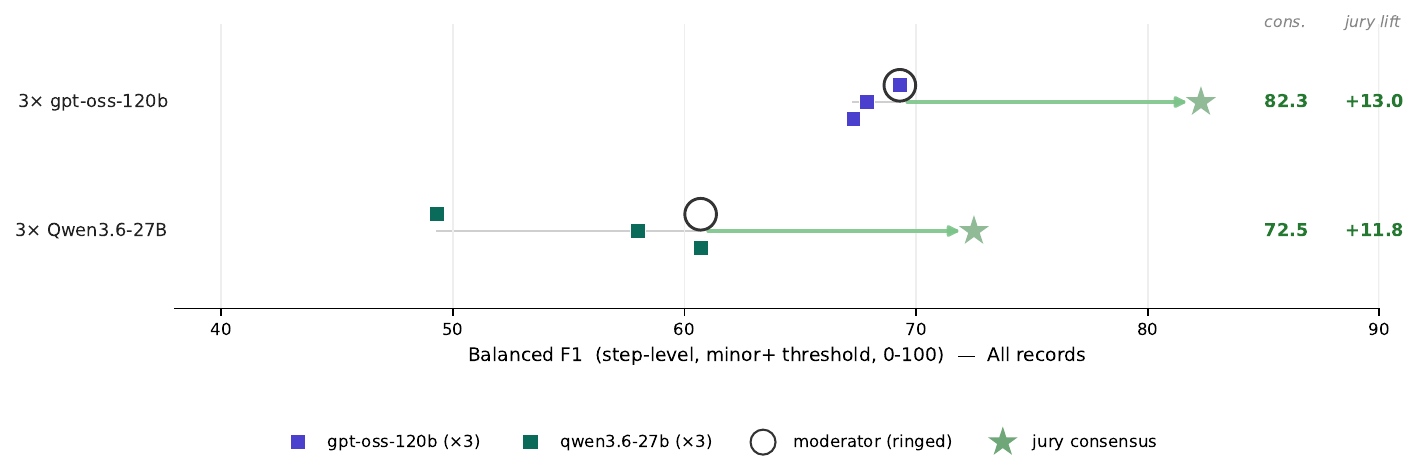}
\caption{Diversity control: homogeneous juries of three independent temperature-$1.0$ samples of one model with a same-model moderator. Jury deliberation lifts consensus over the best single solo even with zero diversity ($+13.0$ for gpt-oss-120b, $+11.8$ for qwen3.6-27b), but the ceiling tracks base-model strength. Exact numbers in \Cref{tab:results-diversity}.}
\label{fig:h2v-diversity}
\end{figure}

We ask whether the jury's gains come from model diversity or simply from aggregating multiple opinions. To isolate this we build homogeneous juries: three independent samples of a single model, moderated by that same model, so that all diversity comes from sampling rather than from mixing architectures (\Cref{fig:h2v-diversity}). A homogeneous gpt-oss-120b jury with deliberation reaches $82.3$ Balanced F1 consensus, a lift of $+13.0$ against the best solo score and significantly higher than opus-4.6, sonnet-4.6, and gemini-3.1-pro. A homogeneous qwen3.6-27b jury reaches $72.5$ against $60.7$ solo, a lift of $+11.8$. In both cases precision stays near-constant and the lift is driven by recall, consistent with the pattern in \Cref{sec:results:jury}.

These results indicate that model diversity is not necessary for large deliberation gains and, in this setting, matters less than model capability. The homogeneous gpt-oss jury reaches $82.3$ Balanced F1, within the roughly $80$ to $84$ range of the diverse juries. In contrast, the homogeneous qwen3.6-27b jury reaches only $72.5$, despite achieving a comparable improvement over its solo baseline. Thus, deliberation across independent samples can be effective without mixing models, but the final performance remains strongly constrained by the capability of the underlying model. These results do not rule out additional benefits from model diversity.

\subsection{Generalization to DeltaBench}\label{sec:results:deltabench}

\begin{figure}[h!]
\centering
\includegraphics[width=\linewidth]{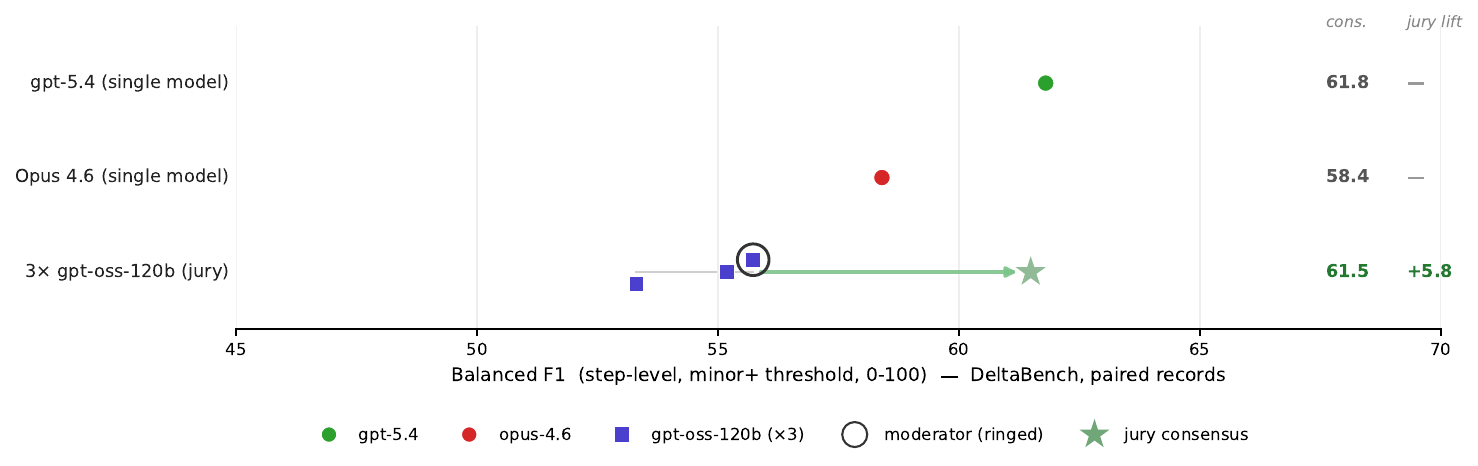}
\caption{DeltaBench generalization: Balanced F1 for a single opus-4.6 Phase 1 pass versus a homogeneous $3\times$ gpt-oss-120b jury with deliberation. Every individual gpt-oss-120b juror trails opus by around $3$--$5$ points, but deliberation lifts the panel $+5.8$ over its best member and past opus-4.6 and at the same level as gpt-5.4. Under DeltaBench's own macro-F1 protocol (see \Cref{app:deltabench}) the jury deliberation scores $47.9$ vs.\ $43.4$ for opus-4.6 (\Cref{app:deltabench}).}
\label{fig:deltabench-strip}
\end{figure}

In this section we evaluate Reasoning Jury on DeltaBench \citep{he-etal-2025-large}, which includes long reasoning traces with human-annotated erroneous steps. We compare a homogeneous $3\times$ gpt-oss-120b jury against a single opus-4.6 Phase 1 pass, on the $1{,}226$ records. For evaluation details refer to \Cref{app:deltabench}.
\Cref{fig:deltabench-strip} shows that the jury's deliberated consensus reaches $61.5$ Balanced F1 against $58.4$ for opus-4.6, with each individual gpt-oss juror below both ($53.3$--$55.7$), and we see that deliberation lifts the panel $+5.8$ over its best member (full table in \Cref{app:deltabench}).

\subsection{Jury deliberation vs majority vote or union of defects}
\label{subsec:delib-vs-con}

One natural question is what if jury deliberation or consolidation is replaced with something simple such as majority vote or a simple union of defects. \Cref{tab:delib-vs-agg-jury,tab:delib-vs-agg-diversity} report the full metric breakdown for both baselines against deliberation across all six panels. 

Majority voting consistently underperforms because exact-step agreement is sparse. This collapses recall to 38--63\% despite precision holding at 86--93\%, dragging Balanced F1 roughly 9--26 points below consensus. Union aggregation captures most of the benefit of multiple independent judgements. Deliberation outperforms union in four of six panels, but underperforms it in two. Thus, the principal accuracy gain comes from pooling independent judgements, but it is just a bag of possibly-duplicate, possibly-contradictory raw claims with no resolution of severity or evidence quality. Consolidation or deliberation, on the other hand, primarily adjudicates, deduplicates, and structures those judgements rather than uniformly improving step-level Balanced F1. These modes produce a single deduplicated, evidence-backed verdict, which has real downstream applications (\Cref{sec:intro}) independent of any metric. However, since benchmark scoring ignores explanation text, severity distinctions, evidence, and dissent, these results do not establish that consolidation or deliberation improves those properties. A more detailed analysis of this is available in \Cref{app:delib-vs-aggregation}.

\subsection{Jury cost analysis}\label{sec:results:cost}

How much does a deliberating jury cost compared to a single strong model? We compare two configurations that ran on the same $200$ Hard2Verify records: a $3\times$ gpt-oss-120b homogeneous jury against a single opus-4.6 Phase 1 call (one model judging alone). Token counts are taken from each run's logged LLM calls, and pricing is per-million tokens at list rates.\footnote{Pricing used: opus-4.6 \$5.00\,/\,1M input, \$25.00\,/\,1M output. gpt-oss-120b \$0.15\,/\,1M input, \$0.62\,/\,1M output. Source: https://aws.amazon.com/bedrock/pricing/ as of July 2026}

\begin{table}[!htbp]
\begin{minipage}{\linewidth}
\centering
\setlength{\tabcolsep}{4pt}
\captionof{table}{Cost comparison on Hard2Verify: a single opus-4.6 Phase 1 call versus the $3\times$ gpt-oss-120b panel under both Phase 2 modes, consolidation and deliberation. Ratios are relative to the opus column. Deliberation uses $7.6\times$ more tokens than opus yet costs $6.4\times$ less; consolidation cuts the jury's cost roughly in half again (\$$6.50$, $12\times$ cheaper than opus) while giving up $2$ Balanced F1 points of the deliberation lift. Both jury modes are more accurate than the frontier single judge.}
\label{tab:results-cost}
\small
\begin{tabular}{lccccc}
\toprule
& & \multicolumn{2}{c}{$3\times$ gpt-oss, consolidation} & \multicolumn{2}{c}{$3\times$ gpt-oss, deliberation} \\
\cmidrule(lr){3-4} \cmidrule(lr){5-6}
Metric & opus (single) & Value & Ratio & Value & Ratio \\
\midrule
Output tokens & $2{,}905{,}719$ & $9{,}276{,}793$ & $3.2\times$ & $16{,}106{,}489$ & $5.5\times$ \\
Input tokens & $1{,}297{,}866$ & $4{,}975{,}354$ & $3.8\times$ & $15{,}765{,}867$ & $12.1\times$ \\
Total tokens & $4{,}203{,}585$ & $14{,}252{,}147$ & $3.4\times$ & $31{,}872{,}356$ & $7.6\times$ \\
LLM calls & $200$ & $800$ & $4.0\times$ & $2{,}173$ & $10.9\times$ \\
\midrule
Input cost & \$$6.49$ & \$$0.75$ & $0.12\times$ & \$$2.36$ & $0.36\times$ \\
Output cost & \$$72.64$ & \$$5.75$ & $0.08\times$ & \$$9.99$ & $0.14\times$ \\
\textbf{Total cost (200 rec)} & \textbf{\$79.13} & \textbf{\$6.50} & $\mathbf{0.08\times}$ & \textbf{\$12.35} & $\mathbf{0.16\times}$ \\
Cost per record & \$$0.396$ & \$$0.033$ & $0.08\times$ & \$$0.062$ & $0.16\times$ \\
\midrule
Balanced F1 & $73.7$ & $80.3$ & --- & $82.3$ & --- \\
\bottomrule
\end{tabular}
\end{minipage}
\end{table}

\Cref{tab:results-cost} reports the comparison. In deliberation mode, despite using $7.6\times$ more tokens in total and making $10.9\times$ more LLM calls, the three-juror gpt-oss jury costs \textbf{\$12.35} for the full $200$-record run, versus \textbf{\$79.13} for a single opus pass, which is roughly $6.4\times$ cheaper. The per-token price difference (opus-4.6 \$25/M output versus gpt-oss \$0.62/M) more than compensates for the higher token volume. In consolidation mode we see that the cost drops to $\$6.50$. On accuracy the jury with deliberation scores $82.3$ Balanced F1 and with consolidation scores $80.3$ against opus-4.6's $73.7$, and gpt-5.4's $83.9$. Compared to opus-4.6 the jury is both more accurate and substantially cheaper (up to 12$\times$). Compared to gpt-5.4 the performance of the jury is similar while still at a fraction of its cost. The practical implication is that when the jurors are cheap open-weight models, the deliberation overhead is dwarfed by the per-token savings relative to a single expensive frontier judge. Multi-turn deliberation comes with additional latency cost which should also be considered.

% ============================ DEFECT PROFILING ============================
\section{Exploratory jury output profiling}\label{sec:profiling}

The results so far use the jury as an evaluator scored against a benchmark. Because its output is a structured, step-grounded, severity-tagged list of defects, the jury can also be used as an \emph{instrument} to run over many reasoning traces from a given model and aggregate the defects into a distribution over defect kinds. This especially would add a rich layer of analysis and insights on top of performance benchmarks, and would go beyond accuracy numbers to surface detailed failure modes. This section applies Reasoning Jury this way to reasoning traces for AIME2026 \cite{dekoninck2026matharena} generated by nemotron-3-super as a case study, producing a reasoning defect fingerprints over a shared taxonomy of defect types for math problems. Such reasoning defect analysis does require a taxonomy as a pre-requisite. For demonstration of this use case, to achieve this taxonomy, we categorize reasoning defects from different models on this benchmark. Next we map defects from nemotron-3-super to this taxonomy and analyze the results. We should emphasize that this taxonomy is not a comprehensive and validated taxonomy of defects for math reasoning, and it's created simply for demonstration purposes, and it does not establish a standard. Also since the jury natural language verdicts are not validated in our experiments the identified defects cannot be taken at face value.

\subsection{Setup}\label{sec:profiling:setup}

We generate reasoning traces from gpt-oss-120b, qwen3.6-27b, and gemma-4-31b on the AIME2026 benchmark. For each model we draw 32 samples per problem. Every reasoning trace is segmented into \code{[STEP-x]} units with the regular-expression segmenter of \Cref{sec:method:steps}, and Reasoning Jury is run over each trace; the final verdict's defects are the signal we aggregate. The jury (glm-5.2, minimax-m3, deepseek-v4-pro, kimi-k2.6) is \emph{disjoint} from the three generators under study, so no model grades its own traces. This is a use of the jury as a defect detector, not an accuracy measurement against human defect labels. For the purpose of the rest of this section, we then induce a single \emph{emergent, shared} taxonomy over the \emph{pooled} defects of all three models. For details please refer to \Cref{app:defect-taxonomy}.

\subsection{Defect fingerprints}
\label{sec:profiling:fingerprints}

The taxonomy makes it possible to ask, for a given model, \emph{what kind} of reasoning failure it exhibits and how severe each kind tends to be. We illustrate this with a close read of one generator, nemotron-3-super, on AIME2026. Over its $1{,}920$ traces ($295{,}453$ steps, an average of $84.6$ tokens per step and $13{,}025$ tokens per trace), the jury flags $8{,}711$ distinct steps, counting each step once at the highest severity assigned to it: $3{,}681$ at \emph{fatal} severity, $3{,}270$ \emph{major}, $1{,}703$ \emph{minor}, and $57$ \emph{neutral}.

\begin{figure}[t]
\centering
\includegraphics[width=\linewidth]{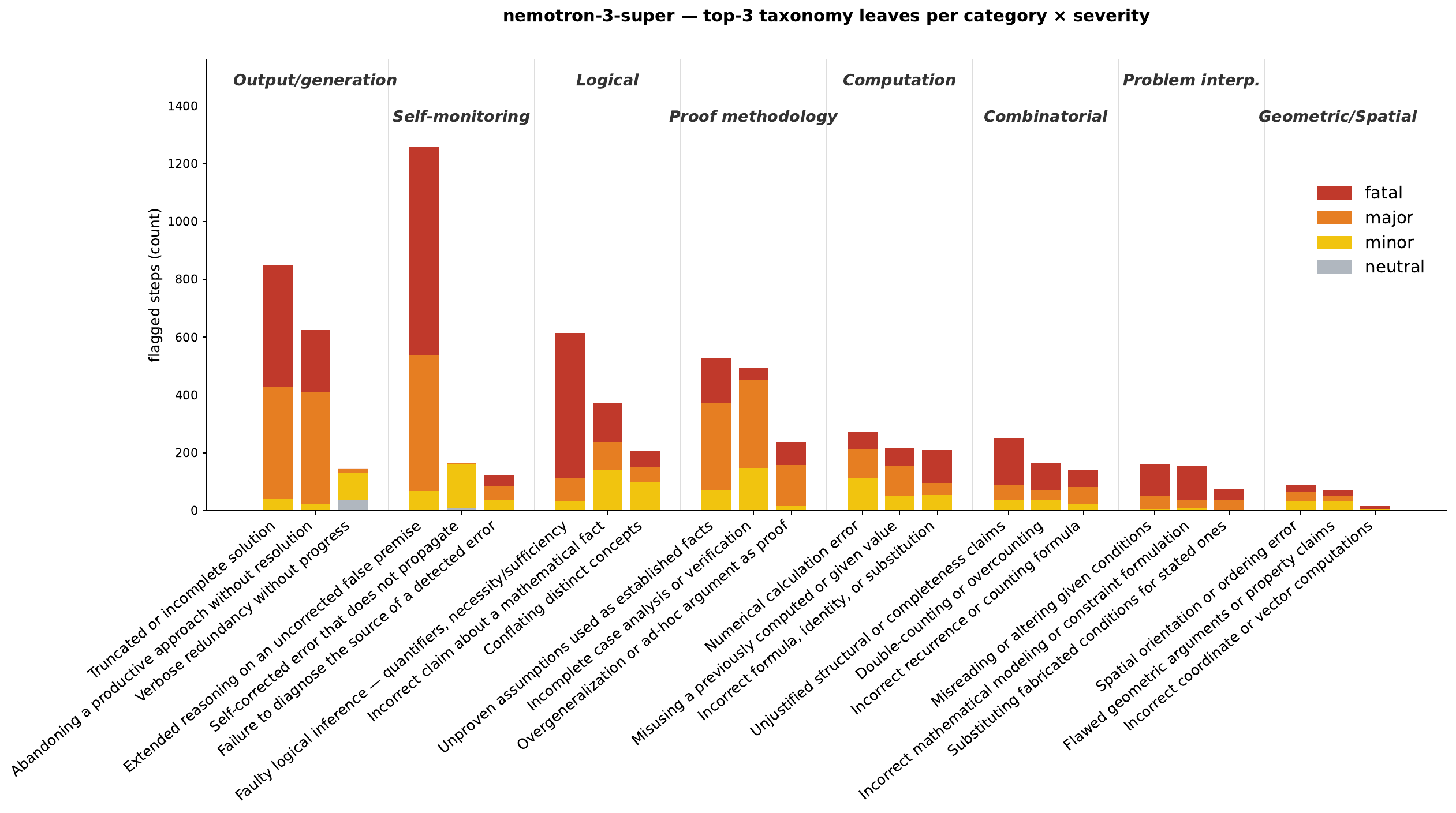}
\caption{nemotron-3-super: flagged steps for the top three taxonomy leaves within each category, stacked by severity (regex segmentation, AIME2026). Within self-monitoring and error-handling, \emph{extended reasoning on an uncorrected false premise} is by far the largest leaf and heavily fatal; within proof methodology and rigor, the dominant leaves (\emph{incomplete case analysis} and \emph{unproven assumptions}) carry far less fatal mass.}
\label{fig:nemotron-stacked-subcat}
\end{figure}

\paragraph{Defective steps by category and severity} \Cref{fig:nemotron-stacked} in the Appendix breaks the flagged steps down by the eight top-level taxonomy categories, each shown as a bar stacked by severity (within a category a step is counted once at its highest severity; a step flagged under two different categories appears in both bars).
\Cref{fig:nemotron-stacked-subcat} drills one level deeper, restricting each category's bar to its top three taxonomy leaves (still stacked by severity), which makes visible which specific defect type is doing the work behind each category's total in \Cref{fig:nemotron-stacked}. Within self-monitoring and error-handling, \emph{extended reasoning on an uncorrected false premise} is both the largest single leaf and almost entirely fatal, confirming it as the dominant driver of that category's fatal mass. Within proof methodology and rigor, by contrast, the leading leaves are \emph{incomplete case analysis or verification} and \emph{unproven assumptions used as established facts}, carrying much less fatal mass than the logical-reasoning and self-monitoring leaves, consistent with that category's lapses-in-rigor profile. The full subcategory (leaf) distribution behind this figure, together with a worked example of a trace carrying defects at every severity level, is given in \Cref{app:profiling:nemotron-subcat}.

\subsection{Correlation between defects and final answer correctness}
\label{sec:profiling:crosscheck}

\begin{figure}[t]
\centering
\begin{minipage}[c]{0.44\linewidth}
\centering
\includegraphics[width=\linewidth]{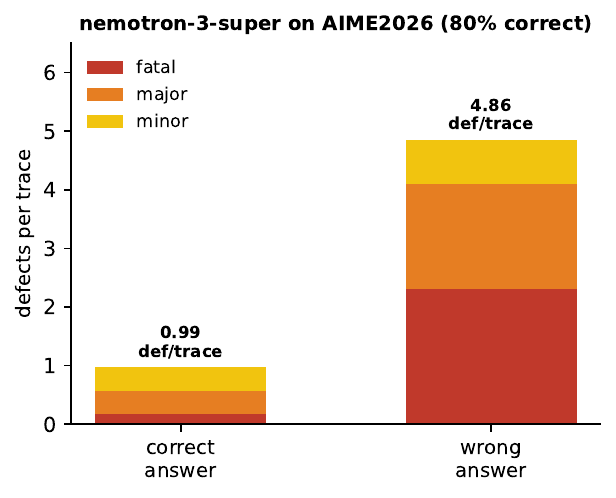}
\end{minipage}\hfill
\begin{minipage}[c]{0.52\linewidth}
\caption{Accuracy vs defect severity for nemotron-3-super on AIME2026 (on which it achieves $80\%$ accuracy). Splitting its traces by correctness of the final answer, the jury flags $0.99$ defects for traces with correct answer, versus $4.86$ for traces with wrong answer, with the share of fatal defects rising from $0.17$ to $2.31$ per trace. The defect signal tracks an external notion of solution quality, and reasoning defects for traces with wrong answer are disproportionately fatal.}
\label{fig:nemotron-crosscheck}
\end{minipage}
\end{figure}

We split nemotron-3-super's traces by whether its final integer answer was correct, and ask whether the jury's defect signal tracks this independent notion of solution quality. If it does, traces with wrong answers should carry both more defects and more \emph{fatal} defects than correct-answer traces from the same model. This is exactly what we observe (\Cref{fig:nemotron-crosscheck}). nemotron-3-super answers $80\%$ (mean@64) of AIME2026 problems correctly; on its correct-answer traces the jury flags $0.99$ defects per trace on average, of which $0.17$ are fatal, while on its wrong-answer traces this rises to $4.86$ defects per trace, of which $2.31$ are fatal, roughly a $5\times$ increase in overall defect rate and a $14\times$ increase in the fatal rate specifically.

% \FloatBarrier

The correct-answer figures are informative in their own right: they are not zero. Even when nemotron-3-super lands on the right final integer, the jury still flags on average just over one defect per trace, exposing ``right answer, flawed reasoning'' traces in which a genuine reasoning defect exists without changing the outcome. The fatal rate on correct traces ($0.17$ per trace) being far from zero but far below the wrong-answer rate ($2.31$) is the expected signature of a defect signal.

\subsection{Caveats}\label{sec:profiling:caveats}

Three caveats bound these fingerprints. First, they are the jury's judgements, not ground truth. Second, the taxonomy is a single induction run at non-zero temperature, so a re-run can produce slightly different labels and counts --- the fingerprints are over \emph{this} taxonomy snapshot. Third, per-step rates are confounded by how finely the segmenter splits each reasoning trace, and as we discussed earlier the approach used for reasoning trace segmentation requires further study.

\section{Exploratory downstream use: defect-guided retry}
\label{sec:profiling:retry}

The final answer correctness correlation above shows that the jury's defect signal tracks
solution correctness, but it does not establish whether information beyond the
flagged step locations helps a model correct its reasoning. We therefore run a
larger, task-based evaluation in which the generating model receives the
jury's feedback and attempts the problem again.

\paragraph{Design.}
We draw $64$ nemotron-3-super samples for each of the $30$ AIME2026
problems, giving $1{,}920$ traces under the same generation settings as
\Cref{sec:profiling:setup}. The original traces achieve $80.0\%$ pass@1. We
evaluate every trace with the same jury, which flags $877$ traces ($45.7\%$)
with at least one non-neutral defect. Among all of the traces, there are $384$ that have wrong final answers. Of these $384$ traces $372$ of them include at least one non-neutral defect, which corresponds to $96.9\%$ recall.

For each flagged trace, the same generator retries the problem under two
matched conditions. In the \emph{step-anchors-only} condition, it receives the
cited \code{[STEP-k]} locations but no description of the defects. In the
\emph{rich-feedback} condition, it receives those same locations together
with the \code{what\_went\_wrong} diagnosis and severity of each defect. Both
conditions include the original problem and trace and instruct the model to
re-solve the problem from scratch. The comparison therefore holds
localization fixed and tests the joint incremental value of diagnosis and
severity.

\begin{center}
\begin{minipage}{\linewidth}
\centering
\small
\setlength{\tabcolsep}{5pt}
\begin{tabular}{@{}lcccc@{}}
\toprule
Feedback & Retry accuracy & Wrong$\to$right & Right$\to$wrong & Overall accuracy \\
\midrule
Step only    & $624/877$ ($71.2\%$) & $146/372$ & $27/505$ & $1{,}655/1{,}920$ ($86.2\%$) \\
Rich defect feedback & $668/877$ ($76.2\%$) & $181/372$ & $18/505$ & $1{,}699/1{,}920$ ($88.5\%$) \\
\bottomrule
\end{tabular}
\captionof{table}{Defect-guided retry on nemotron-3-super AIME2026 traces
($64$ samples per problem). Retry accuracy is measured on the $877$ traces
that have one or more defects with severity more than neutral. Before retry, accuracy is
$505/877$ ($57.6\%$) on the flagged subset and $1{,}536/1{,}920$ ($80.0\%$)
over the full corpus. Steps only is when in a retry the model is given the step indices that include a defect.
In Rich defect feedback the model also receives what went wrong and the severity of each defect for the defective steps.}
\label{tab:retry-main}
\end{minipage}
\end{center}

\paragraph{Results.}
Out of initial $877$ traces with non-neutral defects, $505$ of them ($57.6\%$) initially have a correct answer. Providing rich feedback back to the model for these $877$ problems in a retry produces $668$ correct answers ($76.2\%$), compared with $624$
($71.2\%$) from step anchors alone. Compared to defective steps only information, rich feedback both repairs more of the $372$ flagged
wrong answers ($181$ vs.\ $146$) and causes fewer regressions among the $505$
flagged correct answers ($18$ vs.\ $27$). In the paired comparison, $78$
traces are correct only under rich feedback and $34$ only under step anchors
(exact McNemar $p=4\times10^{-5}$). Because traces from the same problem are
correlated, we also perform a paired analysis at the problem level. The mean
advantage is $+4.1$ percentage points, with a bootstrap $95\%$ confidence
interval of $[+2.3,+6.1]$; rich feedback outperforms step anchors on $15$ of
the $29$ problems with flagged traces, underperforms on one, and ties on $13$.

These results provide task-based evidence for the \emph{joint} value of the
\code{what\_went\_wrong} diagnosis and severity beyond defective step localization alone.
The complete prompts, results, and limitations are in \Cref{app:retry}.

% ============================ LIMITATIONS ============================
\section{Limitations}
\label{sec:limitations}

Reasoning Jury gets its accuracy gains with additional latency and token consumption. Because the jury deliberates, it is inherently slower than a single judge. Phase 1 is fast due to judgements produced in parallel. The consolidation mode also is relatively fast since it is a single LLM call on top of Phase 1. The deliberation mode, however is sequential by design and the number of turns is not known in advance. For offline uses such as data curation or evaluation this added wall-clock time might not introduce a risk. However for online use cases the current deliberation pipeline might not be directly usable for on-policy RL. For off-policy RL where the actor could be a few optimization step behind the policy being optimized, this latency is still tolerable. Additionally, deliberation also consumes far more tokens than a single
pass. Running many jurors, and then having them argue over several turns, means the same trace is read and reasoned about repeatedly. In our cost comparison (\Cref{sec:results:cost}), the jury used roughly $7.6\times$ more tokens and made about $10.9\times$ more LLM calls than a single-model judge on the same records.
As we show there, this does not necessarily translate into higher dollar cost, but still, the raw token and call counts are substantially higher, which matters wherever throughput and rate limits are the binding constraint.

% ============================ CONCLUSION ============================
\section{Conclusion}
\label{sec:conclusion}

We introduced \emph{Reasoning Jury}, a framework for identifying defects in
long reasoning traces with a panel of LLM judges. Jurors first inspect a
step-segmented trace independently; their findings are then combined through
either consolidation or moderated deliberation. The resulting verdict
grounds each defect in one or more \code{[STEP-x]} locations and records what
went wrong, its severity, and supporting evidence. This moves reasoning
evaluation beyond a single score toward feedback that can be inspected,
aggregated, and returned to a model. The benchmark evaluations in this paper
directly measure defect localization; they do not by themselves validate
every field in the structured verdict.

The experiments show that the value of a jury depends on the strength and
composition of its members. On Hard2Verify, the full Frontier jury is already
saturated by gpt-5.4, reaching $84.4$ Balanced F1 compared with $83.9$ for
gpt-5.4 alone. When no single juror dominates, however, pooling independent
judgements substantially improves recall: the two open-weight juries reach
$80.2$--$80.8$ Balanced F1, roughly $11$--$16$ points above their best
individual members. Homogeneous juries also improve over repeated samples of
the same model, showing that independent sampling contributes even without
model diversity. The DeltaBench results exhibit the same pattern, with a
three-member gpt-oss-120b jury reaching $61.5$ Balanced F1, compared with
$58.4$ for opus-4.6 and $61.8$ for gpt-5.4.

The aggregation controls refine the interpretation of these gains. Majority
voting performs poorly because jurors rarely identify exactly the same steps,
whereas the union of their findings captures much of the improvement.
Deliberation does not uniformly outperform union on step-level Balanced F1;
its role is instead to adjudicate conflicting claims, remove duplicates, and
produce one structured verdict. This distinction matters operationally. On
Hard2Verify, a homogeneous gpt-oss-120b jury scores $82.3$ with deliberation
and $80.3$ with consolidation, compared with $73.7$ for a single opus-4.6
judge. Despite using more calls and tokens, the two jury configurations cost
\$12.35 and \$6.50, respectively, versus \$79.13 for opus-4.6 at the list
prices used in our analysis.

The structured findings also support uses that step-level benchmark scores do
not capture. In the AIME2026 case study, the number and severity of detected
defects track final-answer correctness and expose flawed reasoning even in
some correct-answer traces. More directly, when the generating model retries
flagged traces, providing the jury's diagnosis and severity raises accuracy
from $71.2\%$ with step locations alone to $76.2\%$. This result provides
initial task-based evidence that information beyond localization is useful,
although it evaluates diagnosis and severity jointly and does not establish
the accuracy of every individual finding.

Taken together, the results position Reasoning Jury as a practical approach
for offline reasoning evaluation when a single judge is insufficient and a
structured verdict is more useful than a scalar score. Independent sampling
provides the main accuracy benefit, while consolidation and deliberation turn
the pooled findings into a usable output with different cost--latency
tradeoffs. Future work should directly evaluate the factuality and usefulness
of the defect descriptions, calibrate severity labels, isolate the effect of
deliberation under matched compute, and test downstream use across additional
models and domains.

\bibliographystyle{plainnat}
\bibliography{references}

\appendix
% ============================ APPENDIX ============================
\section{Defect schema and an illustrative example}
\label{app:schema}

Each Phase 1 \emph{negative} is a JSON object grounded to specific steps. The
prompt enforces a genericness test and a 1--5 specificity self-check, keeping only
issues scoring $\geq 4$. The contrast below (both abridged from the prompt
templates) shows the bar.

\begin{lstlisting}[caption={Insufficient (specificity score 2 --- rejected).},captionpos=b]
{
  "statement_refs": ["STEP-6", "STEP-14"],
  "what_went_wrong": "The reasoning makes unsupported assumptions about the
     connection between the two domains and applies formulas incorrectly.",
  "impact": "major",
  "evidence": "The trace assumes overlap without justification."
}
\end{lstlisting}

\begin{lstlisting}[caption={Sufficient (specificity score 5 --- accepted).},captionpos=b]
{
  "statement_refs": ["STEP-14", "STEP-17"],
  "what_went_wrong": "In [STEP-14], the trace states 'Energy required is
     proportional to 1/sqrt(final particle size)' citing Rittinger's Law, but
     Rittinger's Law is E proportional to (1/D2 - 1/D1), i.e. 1/D not 1/sqrt(D).
     This misformulation propagates to [STEP-17], yielding an energy estimate
     ~4.5x too low.",
  "impact": "fatal",
  "evidence": "[STEP-14] 'Energy ... (proportional to 1/sqrt(final particle
     size))'; [STEP-17] 'so energy scales as 1/sqrt(50) ...'",
  "correct_value": "E proportional to (1/D2 - 1/D1). For D1=1mm, D2=50um:
     ~19x baseline, not 4.5x."
}
\end{lstlisting}

\section{Phase 1 prompt}
\label{app:phase1prompt}

The complete Phase 1 independent-judgement prompt is reproduced below: a short
system prompt followed by the user prompt template, whose \code{problem},
\code{reasoning\_trace}, and \code{final\_solution} placeholders are filled per
trace. Mathematical symbols are transcribed to ASCII (e.g.\ ``proportional to''
for $\propto$, ``1/sqrt(D)'' for $1/\sqrt{D}$), matching \Cref{app:schema}.

\begin{lstlisting}[caption={The Phase 1 independent-judgement prompt (system prompt and user template).},captionpos=b]
[SYSTEM PROMPT]
You are a reasoning-trace auditor. You identify concrete weaknesses in reasoning traces by pointing to specific steps. You always return valid JSON and never wrap it in markdown fences.

[USER PROMPT]
You are a **reasoning-trace auditor**. Your job is to **evaluate a specific reasoning trace** for a specific problem by identifying concrete weaknesses.

You will be given:
1) **PROBLEM**: the task statement
2) **REASONING TRACE**: the reasoning produced by a model while attempting the problem. The trace is divided into segments, each prefixed with [STEP-x] where x is an integer indicating the ordinal position of each segment in the trace.
3) **FINAL Response**: the final response to the problem, based on the reasoning trace.

## Primary objective
Judge the *weaknesses* of the provided reasoning trace by pointing to **specific bad reasoning moves**.

## Critical constraints
- **Do NOT write a fresh full response** to the problem.
- **Avoid generic feedback.** Every critique must be **trace-grounded**:
  - name the exact quantity / expression / claim involved, and
  - reference which [STEP-x] it occurs in.

## Grounding requirement (very important)
All judgements must reference specific [STEP-x] markers in the reasoning trace. In your `what_went_wrong` description, you MUST:
1. Name the specific step(s): "In [STEP-14], the trace states..."
2. Quote the exact problematic claim from that step
3. Explain WHY it is wrong, including what the correct value/reasoning should be

## Genericness test (must pass)
For every point you make, ask:
> "Could this comment apply to a totally different problem with no edits?"
If yes, rewrite it so it mentions at least one **problem-specific artifact** (variable name, expression, unit, constraint, theorem/rule, diagram type, etc.) AND a **specific operation/transition** from a specific [STEP-x].

## Specificity self-check (must pass)
Before including each issue, score it 1-5:
  5 = cites exact formula/number/quote from a specific STEP AND explains the correct value
  4 = cites a specific claim from a specific STEP AND names the error type
  3 = references a step but describes the issue generically
  2 = could apply to a different problem with minor edits
  1 = completely generic
**Only include issues scoring >= 4.** Discard issues scoring 3 or below.

## Examples

INSUFFICIENT (score 2 -- do NOT write like this):
```json
{
  "statement_refs": ["STEP-6", "STEP-14"],
  "what_went_wrong": "The reasoning makes unsupported assumptions about the connection between the two domains and applies formulas incorrectly.",
  "impact": "major",
  "evidence": "The trace assumes overlap without justification."
}
```

SUFFICIENT (score 5 -- write like this):
```json
{
  "statement_refs": ["STEP-14", "STEP-17"],
  "what_went_wrong": "In [STEP-14], the trace states 'Energy required proportional to 1/sqrt(final particle size)' citing Rittinger's Law, but Rittinger's Law is E proportional to (1/D2 - 1/D1), proportional to 1/D not 1/sqrt(D). This misformulation propagates to [STEP-17] where the energy estimate is computed using the wrong exponent, yielding a value roughly sqrt(20) ~ 4.5x too low.",
  "impact": "fatal",
  "evidence": "[STEP-14] 'Energy required proportional to surface area generated (proportional to 1/sqrt(final particle size))'; [STEP-17] 'so energy scales as 1/sqrt(50) relative to 1mm baseline'",
  "correct_value": "Rittinger's Law: E proportional to (1/D2 - 1/D1). For D1=1mm, D2=50um: E proportional to (1/0.05 - 1/1) = 19, so ~19x baseline, not 4.5x."
}
```

## Severity guidelines (preliminary)
Your severity rating is a **preliminary estimate**. It will be independently recalibrated by a later verification stage. Focus your energy on **accurate detection and specific description** rather than agonizing over the exact severity level. That said, use this guidance:

- **fatal**: The trace teaches a false generalizable rule or fact -- a large language model trained on this would learn fundamentally broken reasoning that transfers to future tasks. Examples: wrong formula, fabricated evidence, factual inversion, category error. Domain-knowledge errors (wrong facts, wrong formulas) are particularly strong fatal signals.
- **major**: A significant error that weakens the reasoning but is localized -- it does not encode a transferable false rule. Examples: arithmetic mistake in one step, omitting a relevant constraint, misreading one data point.
- **minor**: A small imprecision or stylistic issue that does not affect the reasoning outcome. Examples: rounding differences, informal language, minor notation inconsistency.
- **neutral**: Not actually a flaw -- just a different approach or style choice.

**Important**: Do NOT flag **exposition-only** issues (confusing explanation of correct reasoning, poor formatting, verbose presentation) as major or fatal. These affect communication quality, not reasoning quality, and are almost never true flaws for training purposes.

## Input

### PROBLEM
==== begin problem ====
{problem}
==== end problem ====

### REASONING TRACE (the trace to audit)
==== begin reasoning trace ====
{reasoning_trace}
==== end reasoning trace ====

### Final response
==== begin final response ====
{final_solution}
==== end final response ====

## Output format (JSON ONLY)
Return **only valid JSON**. The JSON must include:
- `negatives`: array of trace-grounded weaknesses, each with:
  - `statement_refs`: array of step IDs where the issue manifests (e.g., ["STEP-14", "STEP-17"])
  - `what_went_wrong`: specific description that embeds [STEP-x] references and direct quotes from the trace. Must name the exact quantity/claim and explain why it is wrong.
  - `impact`: one of "neutral" | "minor" | "major" | "fatal"
  - `evidence`: direct quotes from the trace, prefixed with their step IDs (e.g., "[STEP-14] 'Energy proportional to 1/sqrt(D)'")
  - `correct_value`: what the correct reasoning/value/conclusion should be (omit only if the issue is about missing analysis rather than a wrong claim)

## Additional rules
- Be **surgically specific**: state the exact issue and exact error pattern
- Keep everything grounded in the reasoning trace and problem statement
- The reasoning trace is not the final answer -- it's the internal reasoning process that is used to create the final response. So the reasoning trace does not have to follow all the detailed instructions in the problem such as putting the final answer in `\boxed`, ...
- Each `what_went_wrong` must be self-contained: a reader should understand the issue from that field alone, without needing to read `statement_refs` or `evidence` separately

## Defect propagation
- **Important**: Any step that contains or is based on an error is considered incorrect and needs to be marked as having a defect. That is, if the error is carried forward from a previous step or is based on an error in the previous step, consider the step incorrect and give it a defect in the output. A propagated step inherits the **same severity (`impact`)** as the upstream defect it depends on -- if it builds on a `fatal` error, mark it `fatal`.
\end{lstlisting}

\section{Phase 2 prompts}
\label{app:phase2prompts}

Phase 2 uses two prompts: the moderator prompt that runs the deliberation loop
(\Cref{app:moderatorprompt}) and the juror contribution prompt that each selected
juror answers on its turn (\Cref{app:phase2contribution}).

%%%%%%%%%%%%%%%%%%%%%%%%%%%%

\subsection{Consolidation addendum}
\label{app:consolidationprompt}

In consolidation mode (\Cref{sec:results:cost}) the multi-turn deliberation and
the separate verdict extraction are replaced by a single call: the moderator
model re-performs the Phase 1 task with every juror's independent judgement
attached as unverified candidate findings. The consolidator's user prompt is the
Phase 1 prompt of \Cref{app:phase1prompt} verbatim (including the defect-propagation
addendum when propagation is enabled) with the addendum below appended --- the
\code{jury\_count} and \code{panel\_judgements} placeholders are filled with the
panel size and the formatted Phase 1 judgements, attributed by juror id. Unlike
the deliberation moderator (\Cref{app:moderatorprompt}), which is deliberately
blind to the trace, the consolidator sees the full problem and reasoning trace
and rules on every candidate directly; its JSON output is the final verdict,
with two extra per-finding fields (\code{vote\_count}, \code{voters}) recording
panel support. Mathematical symbols are transcribed to ASCII, matching
\Cref{app:schema}.

\begin{lstlisting}[caption={The consolidation addendum, appended verbatim to the Phase 1 prompt of \Cref{app:phase1prompt}.},captionpos=b]
## PANEL CANDIDATE FINDINGS (unverified -- leads, not conclusions)

Before you, {jury_count} independent auditors examined this same trace under the
exact instructions above. Their raw findings are reproduced below, attributed by
auditor id. Treat them as CANDIDATES, not established facts: auditors can be
wrong, can duplicate one another, and can miss defects entirely.

==== begin panel findings ====
{panel_judgements}
==== end panel findings ====

## How to use the panel findings

1. **Verify before adopting.** For each candidate, re-read the cited [STEP-x]
   in the reasoning trace. Adopt it only if the trace itself supports the
   claim. Reject candidates the trace does not support -- never include a
   finding merely because one or several auditors raised it. Head-count is
   not evidence; only the trace is.
2. **Merge duplicates at maximum specificity.** When several candidates
   describe the same underlying defect, output ONE finding, using the MOST
   SPECIFIC description available among them (exact quotes, numbers, formulas,
   [STEP-x] refs). Supplement, don't average: fold in supporting evidence from
   the other auditors, but never replace a specific claim with a vaguer
   paraphrase.
3. **Fill the gaps.** You are simultaneously auditing the trace yourself under
   the instructions above. Include genuine defects that NO auditor raised.
   Your own findings are held to the same grounding, genericness, and
   specificity requirements as everything else.
4. **Recalibrate severity yourself.** A candidate's severity rating is a
   suggestion, not a constraint: assign severity from your own reading of the
   severity guidelines above.
5. **Same bar for everything.** Every finding in your output -- adopted,
   merged, or newly added -- must pass the genericness test and score >= 4 on
   the specificity self-check.

## Output format reminder (JSON ONLY -- same schema as specified above)

Your final output remains **only valid JSON**, exactly as specified in the
"Output format" section above -- consolidating the panel findings does NOT
change the output schema. Do not output any prose, commentary on the panel,
or markdown fences. The JSON must include:
- `negatives`: array of trace-grounded weaknesses, each with:
  - `statement_refs`: array of step IDs where the issue manifests (e.g., ["STEP-14", "STEP-17"])
  - `what_went_wrong`: specific description that embeds [STEP-x] references and direct quotes from the trace. Must name the exact quantity/claim and explain why it is wrong.
  - `impact`: one of "neutral" | "minor" | "major" | "fatal"
  - `evidence`: direct quotes from the trace, prefixed with their step IDs (e.g., "[STEP-14] 'Energy proportional to 1/sqrt(D)'")
  - `correct_value`: what the correct reasoning/value/conclusion should be (omit only if the issue is about missing analysis rather than a wrong claim)
  - `vote_count`: how many panel auditors' candidates support this finding (0 if it is yours alone)
  - `voters`: array of the supporting auditor ids (e.g., ["jury-0", "jury-2"]; empty array for your own findings)
\end{lstlisting}

%%%%%%%%%%%%%%%%%%%%%%%%%%%%

\subsection{Moderator prompt}
\label{app:moderatorprompt}

The moderator's prompt is reproduced below: a system prompt establishing its
procedural role, followed by the per-turn template whose placeholders
(\code{total\_turns}, \code{recent\_transcript}, \code{previous\_consensus\_state},
etc.) are filled each turn. The moderator never receives the problem, the
reasoning trace, or the solution; it sees only the deliberation transcript.
Mathematical symbols are transcribed to ASCII, matching \Cref{app:schema}.

\begin{lstlisting}[caption={The moderator prompt (system prompt and per-turn template).},captionpos=b]
[SYSTEM PROMPT]
You are a deliberation moderator for a panel of reasoning-trace auditors.

CRITICAL CONSTRAINTS:
- You have NO access to the problem, reasoning trace, or solution under review.
- You can ONLY see the conversation between panel members.
- You must NEVER speculate about the content of the reasoning trace.
- Your role is purely procedural: manage turns, identify disagreements, and detect convergence.

YOUR RESPONSIBILITIES:

0. CONSENSUS TRACKING: One of your jobs is to produce **one high-fidelity consolidated judgement** by:
   - clustering overlapping findings into canonical, problem-specific items
   - **preserving maximum specificity** from the most detailed source judgement

   ## Critical constraints
   - **Do NOT solve the problem.**
   - **Do NOT invent new issues** not supported by the provided judgements.
   - **Do NOT abstract away specifics.** When multiple judges describe the same issue at different levels of detail, use the MOST SPECIFIC description as the base.

   ## Specificity preservation rules (very important)
   When consolidating overlapping issues into a canonical description:

      1. **Select the best source**: Identify which judge gave the most specific description (most step references, direct quotes, exact numbers/formulas). Use that as the canonical wording.
      2. **Preserve from the best source**:
         - Exact numbers, formulas, or calculations
         - Direct quotes from the trace (keep in quotation marks)
         - Specific [STEP-x] references
         - The `correct_value` if any judge provides it
      3. **Supplement, don't average**: Add supporting evidence from other judges, but never replace a specific claim with a vaguer paraphrase.
      4. **Genericness test**: Each consolidated issue must pass: "Could this description apply to a totally different problem with no edits?" If yes, rewrite using the most specific judge's language.

      Example:
      - jury-2 says: "In [STEP-14], the trace states 'E proportional to 1/sqrt(D)' but Rittinger's Law is E proportional to 1/D"
      - jury-4 says: "The energy formula is applied incorrectly"
      - **CORRECT consolidation**: Use jury-2's wording verbatim. jury-4 adds support but not specificity.
      - **WRONG consolidation**: "The energy formula is misapplied" (lost jury-2's detail)

   ## Output (JSON)
   For this responsibility, return **only valid JSON**, under key "consensus_state" in the final output. The JSON must include:
      - `consensus_map`: array of consensus objects, each with:
      - `what_went_wrong`: consolidated claim -- must be the MOST SPECIFIC version, embedding [STEP-x] refs and direct quotes where available
      - `source_ref`: ["jury-1", "jury-3", ...] -- which judges support this claim
      - `best_source`: "jury-2" -- which judge provided the most specific description used as base
      - `statement_refs`: ["STEP-4", "STEP-7"] -- all [STEP-x] markers involved, collected from the supporting judges' descriptions
      - `consensus_id`: "C1", "C2", etc.
      - `judge_count`: number of judges supporting this issue
      - `impact`: "fatal" | "major" | "minor" | "neutral" -- preliminary severity from the best source judge
      - `evidence`: direct quotes from the trace, prefixed with [STEP-x] -- taken from the most specific judge
      - `correct_value`: what the correct reasoning/value should be (carry from best source if available, omit if no judge provided it)

1. TURN SELECTION: Choose which panelist speaks next based on:
   - Ensure every panelist speaks at least once per logical round.
   - Prioritise panelists involved in unresolved disagreements.
   - Recall silent panelists who have not spoken in 2+ turns.

2. INSTRUCTION GENERATION: Give the selected panelist a specific, process-oriented instruction:
   - Point them to specific disagreements they should address.
   - Ask them to clarify, defend, or concede specific points raised by others.
   - NEVER suggest what the "right" answer is about the trace content.

3. TERMINATION DETECTION: Signal that deliberation should end when:
   - A full logical round passes with no new substantive arguments.
   - All panelists have explicitly signalled agreement on all points.
   - Arguments are cycling without resolution.

You MUST respond with ONLY valid JSON (no markdown fences) in this exact format:
{
  "should_terminate": false,
  "termination_reason": null,
  "next_speaker": "<agent-id>",
  "instruction": "<what you want them to do>",
  "reasoning": "<your internal reasoning about conversation dynamics>",
  "consensus_state": <json from Responsibility 0: CONSENSUS TRACKING>
}

[PER-TURN TEMPLATE]
DELIBERATION STATUS:
- Total turns so far: {total_turns}
- Current logical round: {logical_round}
- Panelists who have spoken this logical round: {spoken_this_round}
- Panelists who have NOT spoken this logical round: {silent_this_round}
- All panelist IDs: {all_agent_ids}

TRANSCRIPT (last {recent_window} turns):

==== begin transcript ====
{recent_transcript}
==== end transcript ====

YOUR PREVIOUS CONSENSUS STATE:
This is the consensus_state you produced on your last turn. Use it as your starting point -- update it based on what has changed in the transcript since then (new endorsements, concessions, withdrawn claims, etc.). On the first turn this will be empty. Keep in mind that the panelists do NOT have access to consensus state, so do not refer to its contents when you address the panelists.

==== begin previous consensus state ====
{previous_consensus_state}
==== end previous consensus state ====

Based on the deliberation dynamics, decide:
0. Update "consensus_state": Pay special attention to the last juror's response to your question and check whether the current speaker endorsed, contested, or was silent on each consensus item. Update `source_ref` on consensus_state accordingly.
1. Should deliberation terminate?  (Has a full round passed with no new substantive arguments? Are panelists repeating themselves?)
2. If not, who should speak next and what should they address? Do NOT refer to anything from consensus state when you address the panelists, as they don't have access to it.
\end{lstlisting}

\subsection{Juror contribution prompt}
\label{app:phase2contribution}

When a juror is selected to speak, it receives the Phase 2 contribution prompt
below. The prompt opens by identifying the juror, then re-includes the same
auditor preamble, problem/trace/solution input, and additional rules as the
Phase 1 prompt (\Cref{app:phase1prompt}), and closes with the deliberation
transcript so far, the moderator's instruction for this turn, and the
critical rules governing the contribution.

\begin{lstlisting}[caption={The Phase 2 juror contribution prompt. The middle blocks are identical to the Phase 1 prompt (\Cref{app:phase1prompt}) and elided here.},captionpos=b]
You are {agent_id}. <<< then the auditor preamble, the PROBLEM / REASONING
TRACE / FINAL RESPONSE input block, and the "Additional rules" block, all
identical to the Phase 1 prompt in Appendix B (the JSON output-format block is
omitted, since a Phase 2 turn is a natural-language argument, not a JSON list) >>>

---

Remember that you are {agent_id}. Your prior contributions in the deliberations so far are in the transcript below (marked by {agent_id}).

DELIBERATION SO FAR:
==== begin deliberation so far ====
{transcript}
==== end deliberation so far ====

---

Now the moderator in the panel has the following instruction for you:

MODERATOR INSTRUCTION FOR YOU:

==== begin moderator instructions for you ====
{instruction}
==== end moderator instructions for you ====

---

Follow the instructions from the moderator with the following rules in mind:

CRITICAL RULES:
- Every claim you make MUST reference specific [STEP-x] markers from the reasoning trace provided above under ==== begin reasoning trace ====. Carefully examine it before making decisions.
- If you cannot ground a claim in a specific step, do not make it.
- You may AGREE with other panelists, DISAGREE with evidence, or RAISE new issues.
- Be concise. Focus on the strongest arguments.
- If you have nothing new to add, say "I have no new arguments to present."

Respond with your contribution to the deliberation.
\end{lstlisting}

\section{Severity rubric}
\label{app:severity}

Severity is assigned on a four-level ordinal scale (neutral $=0$, minor $=1$,
major $=2$, fatal $=3$), with the training-impact framing used in the prompts:

\begin{description}[leftmargin=0pt,itemsep=2pt,topsep=2pt,style=nextline]
  \item[fatal] The trace teaches a \emph{false generalizable rule or fact}---a
    model trained on it would learn broken reasoning that transfers (wrong
    formula, fabricated evidence, factual inversion, category error).
  \item[major] A significant but \emph{localized} error that does not encode a
    transferable false rule (a single arithmetic slip, an omitted constraint, a
    misread data point).
  \item[minor] A small imprecision not affecting the outcome (rounding, notation).
  \item[neutral] Not a flaw---a different valid approach or style choice.
\end{description}
Exposition-only issues (confusing but correct explanations, formatting) are
explicitly \emph{not} to be rated major/fatal: they affect communication, not
reasoning quality.

\section{Verdict object}
\label{app:verdict}

The pipeline emits a \code{Verdict} containing: the consensus \code{negatives} (each
with \code{statement\_refs}, \code{what\_went\_wrong}, \code{impact},
\code{evidence}, \code{vote\_count}, \code{voters}); a scalar \code{confidence}
$\in[0,1]$; \code{dissenting\_views}; \code{deliberation\_rounds};
\code{termination\_reason}; the full \code{transcript} and \code{phase1\_judgements};
and, when the corresponding mechanisms are enabled, \code{dropped\_negatives}
(removal pass) and \code{negative\_adjudications} (evidence arbiter). A
best-effort fallback verdict is produced if extraction fails, so the pipeline
degrades gracefully rather than crashing.

\section{Fault tolerance}
\label{app:faults}

At jury scale, individual calls fail; the pipeline is designed to degrade rather
than crash. If some Phase 1 jurors fail, it proceeds with the remainder above a
configured minimum. A juror failing during deliberation has its turn skipped and
is removed after repeated consecutive failures. A moderator failure falls back to
deterministic round-robin turn order; a grounding-checker failure accepts the
contribution (benefit of the doubt); an extraction failure returns a
best-effort verdict aggregated from Phase 1. API throttling is retried with
exponential backoff, and a Phase 1 wall-clock ceiling prevents one hung juror from
stalling the fan-out.

\section{Full results tables}
\label{app:fulltables}

This section collects the per-configuration numbers behind the figures in \Cref{sec:results}. All numbers are step-level, All-records, minor+, propagation-aware Balanced F1 on a $0$--$100$ scale, scored against the Hard2Verify human labels. \Cref{tab:results-jury-solo} corresponds to \Cref{fig:h2v-jury} (each jury's consensus against its jurors' solo scores), \Cref{tab:results-knockout} to \Cref{fig:h2v-knockout} (jury knockout), \Cref{tab:results-moderator} to \Cref{fig:h2v-modsweep} (moderator sweep), and \Cref{tab:results-diversity} to \Cref{fig:h2v-diversity} (diversity control). The four-panel jury consensus is also reported inline as \Cref{tab:results-jury}.

\begin{table}[t]
\centering
\caption{Per-juror solo Phase 1 versus deliberated consensus for the four juries of \Cref{fig:h2v-jury} (minor+, All records, step-level, propagation-aware). Within each panel the consensus row is the deliberated final verdict; the remaining rows are that panel's jurors judging alone, sorted by Balanced F1. The Bal-F1 column carries a $95\%$ bootstrap confidence half-width ($\pm$, $10{,}000$ record-level resamples).}
\label{tab:results-jury-solo}
\setlength{\tabcolsep}{4pt}
\begin{tabular}{llcccccc}
\toprule
Jury & Member & BalAcc & Bal-F1 & Acc & P & R & F1 \\
\midrule
\multirow{6}{*}{Frontier (5j)}
 & \textbf{Deliberation} & $84.5$ & $\mathbf{84.4}$\,{\scriptsize$\pm2.5$} & $84.1$ & $77.5$ & $87.1$ & $82.0$ \\
  & \textbf{Consolidation} & $83.2$ & $\mathbf{83.2}$\,{\scriptsize$\pm2.7$} & $83.3$ & $78.7$ & $82.3$ & $80.5$ \\
 & gpt-5.4 & $83.9$ & $83.9$\,{\scriptsize$\pm2.8$} & $84.1$ & $78.8$ & $82.8$ & $80.7$ \\
 & opus-4.6 & $76.2$ & $73.7$\,{\scriptsize$\pm3.7$} & $78.4$ & $81.7$ & $62.5$ & $70.8$ \\
 & glm-5.2 & $74.0$ & $69.4$\,{\scriptsize$\pm4.4$} & $77.0$ & $84.1$ & $55.5$ & $66.9$ \\
 & sonnet-4.6 & $74.2$ & $71.2$\,{\scriptsize$\pm3.4$} & $76.6$ & $79.4$ & $59.4$ & $67.9$ \\
 & gemini-3.1-pro & $73.4$ & $69.5$\,{\scriptsize$\pm3.7$} & $76.2$ & $80.8$ & $56.5$ & $66.5$ \\
\midrule
% \multirow{5}{*}{Frontier $-$gpt-5.4 (4j)}
 Frontier $-$gpt-5.4 (4j) & \textbf{Deliberation} & $81.8$ & $\mathbf{81.7}$\,{\scriptsize$\pm2.6$} & $82.3$ & $78.9$ & $78.2$ & $78.8$ \\
 & \textbf{Consolidation} & $81.4$ & $\mathbf{81.3}$\,{\scriptsize$\pm2.7$} & $82.0$ & $78.8$ & $78.0$ & $78.4$ \\
 % & opus-4.6 & $76.2$ & $73.7$\,{\scriptsize$\pm3.7$} & $78.4$ & $81.7$ & $62.5$ & $70.8$ \\
 % & glm-5.2 & $76.6$ & $72.8$\,{\scriptsize$\pm4.4$} & $79.3$ & $87.0$ & $59.6$ & $70.8$ \\
 % & sonnet-4.6 & $74.1$ & $71.2$\,{\scriptsize$\pm3.5$} & $76.5$ & $79.4$ & $59.4$ & $67.9$ \\
 % & gemini-3.1-pro & $73.4$ & $69.5$\,{\scriptsize$\pm3.6$} & $76.2$ & $80.8$ & $56.5$ & $66.5$ \\
\midrule
\multirow{6}{*}{Large OSS (5j)}
 & \textbf{Deliberation} & $80.6$ & $\mathbf{80.2}$\,{\scriptsize$\pm3.1$} & $81.5$ & $79.6$ & $75.0$ & $77.2$ \\
 & \textbf{Consolidation} & $79.7$ & $\mathbf{78.8}$\,{\scriptsize$\pm3.1$} & $81.1$ & $81.1$ & $71.3$ & $75.9$ \\
 & glm-5.2 & $74.0$ & $69.4$\,{\scriptsize$\pm4.4$} & $77.0$ & $84.1$ & $55.5$ & $66.9$ \\
 & deepseek-v4-pro & $72.6$ & $66.5$\,{\scriptsize$\pm4.1$} & $76.0$ & $85.1$ & $51.6$ & $64.3$ \\
 & nemotron-3-super & $64.8$ & $50.6$\,{\scriptsize$\pm6.0$} & $69.7$ & $83.5$ & $34.5$ & $48.8$ \\
 & kimi-k2.6 & $66.7$ & $53.8$\,{\scriptsize$\pm6.7$} & $71.5$ & $87.0$ & $37.4$ & $52.3$ \\
 & minimax-m3 & $64.3$ & $50.4$\,{\scriptsize$\pm6.8$} & $69.2$ & $80.9$ & $34.4$ & $48.3$ \\
\midrule
\multirow{6}{*}{Small/Medium OSS (5j)}
 & \textbf{Deliberation} & $81.0$ & $\mathbf{80.8}$\,{\scriptsize$\pm3.0$} & $81.6$ & $78.3$ & $77.4$ & $77.9$ \\
 & \textbf{Consolidation} & $80.3$ & $\mathbf{79.3}$\,{\scriptsize$\pm3.2$} & $81.8$ & $82.9$ & $71.3$ & $76.6$ \\
 & gpt-oss-120b & $71.9$ & $65.4$\,{\scriptsize$\pm4.5$} & $75.4$ & $85.0$ & $50.3$ & $63.2$ \\
 & minimax-m2.7 & $66.2$ & $58.6$\,{\scriptsize$\pm4.6$} & $69.8$ & $73.4$ & $43.8$ & $54.9$ \\
 & qwen3.6-27b & $67.7$ & $56.3$\,{\scriptsize$\pm5.0$} & $72.2$ & $86.6$ & $39.9$ & $54.6$ \\
 & gemma-4-31b & $66.3$ & $51.3$\,{\scriptsize$\pm6.6$} & $71.4$ & $92.2$ & $34.7$ & $50.5$ \\
 & nemotron-3-super & $64.8$ & $50.6$\,{\scriptsize$\pm6.0$} & $69.7$ & $83.5$ & $34.5$ & $48.8$ \\
\bottomrule
\end{tabular}
\end{table}

\begin{table}[t]
\centering
\caption{Knockout: consensus scores (minor+, All records, step-level, propagation-aware) as each jury is reduced from five to two jurors, with Phase 1 held fixed (each smaller jury reuses the full jury's Phase 1 judgements; only Phase 2 is re-run). Quality holds down to three jurors on every panel; the sharpest drop is at two.}
\label{tab:results-knockout}
\setlength{\tabcolsep}{4pt}
\begin{tabular}{llcccccc}
\toprule
Panel & Size & BalAcc & Bal-F1 & Acc & P & R & F1 \\
\midrule
\multirow{4}{*}{Frontier}
 & 5j & $84.5$ & $84.4$ & $84.1$ & $77.5$ & $87.1$ & $82.0$ \\
 & 4j & $84.6$ & $84.5$ & $84.1$ & $76.7$ & $87.3$ & $81.6$ \\
 & 3j & $84.6$ & $84.5$ & $84.2$ & $76.8$ & $86.7$ & $81.5$ \\
 & 2j & $84.7$ & $84.7$ & $84.4$ & $77.8$ & $85.8$ & $81.6$ \\
\midrule
\multirow{3}{*}{Frontier $-$gpt-5.4}
 & 4j & $81.8$ & $81.7$ & $82.3$ & $78.9$ & $78.8$ & $78.8$ \\
 & 3j & $80.7$ & $80.5$ & $81.4$ & $78.5$ & $76.6$ & $77.5$ \\
 & 2j & $80.5$ & $80.1$ & $81.4$ & $79.4$ & $74.9$ & $77.1$ \\
\midrule
\multirow{4}{*}{Large OSS}
 & 5j & $80.6$ & $80.2$ & $81.5$ & $79.6$ & $75.0$ & $77.2$ \\
 & 4j & $81.8$ & $81.3$ & $82.8$ & $81.6$ & $75.8$ & $78.6$ \\
 & 3j & $79.7$ & $78.9$ & $81.1$ & $81.3$ & $71.3$ & $75.9$ \\
 & 2j & $78.5$ & $77.1$ & $80.2$ & $81.6$ & $67.9$ & $74.2$ \\
\midrule
\multirow{4}{*}{Small/Medium OSS}
 & 5j & $81.0$ & $80.8$ & $81.6$ & $78.3$ & $77.4$ & $77.9$ \\
 & 4j & $80.1$ & $79.7$ & $81.1$ & $79.6$ & $74.0$ & $76.7$ \\
 & 3j & $77.6$ & $76.8$ & $78.9$ & $77.7$ & $69.6$ & $73.4$ \\
 & 2j & $75.4$ & $73.5$ & $77.3$ & $78.3$ & $63.5$ & $70.1$ \\
\bottomrule
\end{tabular}
\end{table}

\begin{table}[t]
\centering
\caption{Moderator sweep with the jury and Phase 1 held fixed: deliberation scores as only the moderator/verdict model varies. Moderator choice changes Balanced F1 by $2.6$--$3.7$ points. Consolidation results are shown in \Cref{fig:h2v-modsweep}.}
\label{tab:results-moderator}
\setlength{\tabcolsep}{4pt}
\begin{tabular}{llcccccc}
\toprule
Panel & Moderator & BalAcc & Bal-F1 & Acc & P & R & F1 \\
\midrule
\multirow{5}{*}{Small/Medium}
 & gemma-4-31b & $82.4$ & $82.0$ & $83.3$ & $82.3$ & $76.8$ & $79.4$ \\
 & gpt-oss-120b (baseline) & $81.0$ & $80.8$ & $81.6$ & $78.3$ & $77.4$ & $77.9$ \\
 & qwen3.6-27b & $81.1$ & $80.5$ & $82.3$ & $81.9$ & $74.1$ & $77.8$ \\
 & minimax-m2.7 & $80.4$ & $80.2$ & $81.0$ & $77.8$ & $76.5$ & $77.2$ \\
 & nemotron-3-super & $79.9$ & $79.4$ & $80.9$ & $79.3$ & $73.7$ & $76.4$ \\
\midrule
\multirow{5}{*}{Large-OSS}
 & minimax-m3 & $82.5$ & $82.3$ & $83.1$ & $80.7$ & $78.4$ & $79.5$ \\
 & glm-5.2 & $81.8$ & $81.7$ & $82.5$ & $79.7$ & $77.9$ & $78.8$ \\
 & kimi-k2.6 & $81.5$ & $81.4$ & $82.1$ & $78.7$ & $78.1$ & $78.4$ \\
 & deepseek-v4-pro (baseline) & $80.6$ & $80.2$ & $81.5$ & $79.6$ & $75.0$ & $77.2$ \\
 & nemotron-3-super & $79.0$ & $78.6$ & $80.0$ & $77.7$ & $73.1$ & $75.3$ \\
\bottomrule
\end{tabular}
\end{table}

\FloatBarrier
  \noindent\begin{minipage}{\linewidth}
  \centering
  \captionof{table}{Diversity control: homogeneous panels of three independent temperature-$1.0$ samples of one model with a same-model moderator (minor+, All records, step-
  level, propagation-aware; $n=200$). Within each panel: the deliberated consensus followed by the three individual solo samples (interchangeable draws of the same model, so
  their spread reflects sampling variability). Even with zero model diversity the consensus rises well above every solo sample, and the gain is recall-driven (precision roughly
  flat or slightly lower), but the ceiling tracks base-model strength.}
  \label{tab:results-diversity}
  \setlength{\tabcolsep}{4pt}
  \begin{tabular}{llcccccc}
  \toprule
  Panel & Configuration & BalAcc & Bal-F1 & Acc & P & R & F1 \\
  \midrule
  \multirow{4}{*}{$3\times$ gpt-oss-120b}
   & \textbf{Deliberation} & $\mathbf{82.5}$ & $\mathbf{82.3}$ & $\mathbf{83.2}$ & $80.8$ & $\mathbf{78.6}$ & $\mathbf{79.7}$ \\
   & Solo juror 1 & $73.2$ & $67.3$ & $76.6$ & $86.6$ & $52.3$ & $65.2$ \\
   & Solo juror 2 & $73.2$ & $67.9$ & $76.4$ & $84.6$ & $53.5$ & $65.5$ \\
   & Solo juror 3 & $74.1$ & $69.3$ & $77.2$ & $85.0$ & $55.3$ & $67.0$ \\
  \midrule
  \multirow{4}{*}{$3\times$ qwen3.6-27b}
   & \textbf{Deliberation} & $\mathbf{75.6}$ & $\mathbf{72.5}$ & $\mathbf{78.1}$ & $82.9$ & $\mathbf{60.3}$ & $\mathbf{69.8}$ \\
   & Solo juror 1 & $70.0$ & $60.7$ & $74.1$ & $87.8$ & $44.5$ & $59.1$ \\
   & Solo juror 2 & $68.0$ & $58.0$ & $72.2$ & $83.6$ & $41.9$ & $55.8$ \\
   & Solo juror 3 & $63.4$ & $49.3$ & $68.3$ & $78.6$ & $33.5$ & $46.9$ \\
  \bottomrule
  \end{tabular}
  \end{minipage}

\section{Deliberation vs.\ naive aggregation: full metrics}
\label{app:delib-vs-aggregation}

\Cref{tab:delib-vs-agg-jury,tab:delib-vs-agg-diversity} give the full metric
breakdown (Bal.~Acc, Bal.~F1, Acc., Precision, Recall, F1) underlying the
comparison between the moderated \textsc{consensus} verdict and two naive
Phase 1 aggregation baselines: the \textsc{union} of every juror's independently
flagged steps, and \textsc{majority vote} (a step counts as flagged only if a
strict majority of jurors --- more than half --- independently cited it). All
rows use the \code{minor+} threshold (positives = minor/major/fatal) on the
\emph{All records} slice, scored step-level against Hard2Verify human labels.

\begin{table}[h]
\centering
\small
\renewcommand{\arraystretch}{1.1}
\setlength{\tabcolsep}{4.5pt}
\begin{tabular}{@{}l rrrrrr@{}}
\toprule
\textbf{Configuration} & \textbf{Bal.~Acc} & \textbf{Bal.~F1} & \textbf{Acc.} &
  \textbf{Prec.} & \textbf{Recall} & \textbf{F1} \\
\midrule
\multicolumn{7}{@{}l}{\emph{Frontier (5 jurors)}}\\
\textbf{Deliberation}    & \textbf{84.5} & \textbf{84.4} & \textbf{84.1} & 77.5 & 87.1 & \textbf{82.0} \\
\quad Union            & 82.0 & 81.5 & 80.9 & 72.2 & \textbf{88.4} & 79.5 \\
\quad Majority vote    & 77.8 & 75.0 & 80.2 & \textbf{85.9} & 63.0 & 72.7 \\
\midrule
\multicolumn{7}{@{}l}{\emph{Frontier $-$gpt-5.4 (4 jurors)}}\\
\textbf{Deliberation}    & 81.8 & 81.7 & \textbf{82.3} & 78.9 & 78.2 & 78.8 \\
\quad Union            & \textbf{82.0} & \textbf{82.0} & 81.9 & 76.2 & \textbf{82.7} & \textbf{79.3} \\
\quad Majority vote    & 75.4 & 70.5 & 78.5 & \textbf{88.1} & 56.3 & 68.7 \\
\midrule
\multicolumn{7}{@{}l}{\emph{Large OSS (5 jurors)}}\\
\textbf{Deliberation}    & \textbf{80.6} & \textbf{80.2} & \textbf{81.5} & 79.6 & 75.0 & \textbf{77.2} \\
\quad Union            & 79.6 & 79.3 & 80.3 & 77.0 & \textbf{75.3} & 76.1 \\
\quad Majority vote    & 69.6 & 59.3 & 74.0 & \textbf{89.2} & 42.8 & 57.9 \\
\midrule
\multicolumn{7}{@{}l}{\emph{Small/Medium OSS (5 jurors)}}\\
\textbf{Deliberation}    & \textbf{81.0} & \textbf{80.8} & \textbf{81.6} & 78.3 & 77.4 & \textbf{77.9} \\
\quad Union            & 79.4 & 79.4 & 79.5 & 74.2 & \textbf{78.5} & 76.3 \\
\quad Majority vote    & 68.0 & 54.9 & 72.8 & \textbf{92.5} & 38.2 & 54.1 \\
\bottomrule
\end{tabular}
\caption{Moderated deliberation vs.\ union and majority-vote aggregation of
independent Phase 1 juror judgements, for the four heterogeneous panels in the
\emph{Jury} strip figure (\Cref{sec:results:jury} main text). Majority vote collapses
recall (38--63\%) despite the highest precision of the three aggregation rules
on every panel, dragging Bal.~F1 well below both union and consensus. Union
recovers most of consensus's Bal.~F1 (within 1--3 points on three of four
panels) but does so via a different precision/recall trade-off in every panel:
on Frontier, union's recall exceeds consensus's (0.884 vs.\ 0.871) at a steep
precision cost (0.722 vs.\ 0.775); on Large~OSS and Small/Medium~OSS, union's
recall is comparable to or below consensus's while precision trails by
2.6--4.1 points. Deliberation's bold entry per panel marks the better of
consensus vs.\ union on each column.}
\label{tab:delib-vs-agg-jury}
\end{table}

\begin{table}[h]
\centering
\small
\renewcommand{\arraystretch}{1.1}
\setlength{\tabcolsep}{4.5pt}
\begin{tabular}{@{}l rrrrrr@{}}
\toprule
\textbf{Configuration} & \textbf{Bal.~Acc} & \textbf{Bal.~F1} & \textbf{Acc.} &
  \textbf{Prec.} & \textbf{Recall} & \textbf{F1} \\
\midrule
\multicolumn{7}{@{}l}{\emph{Homogeneous 3$\times$ gpt-oss-120b ($n{=}200$)}}\\
\textbf{Deliberation}    & \textbf{82.5} & \textbf{82.3} & \textbf{83.2} & \textbf{80.8} & \textbf{78.6} & \textbf{79.7} \\
\quad Union            & 80.7 & 80.2 & 81.8 & \textbf{80.8} & 74.1 & 77.3 \\
\quad Majority vote    & 73.4 & 67.4 & 76.8 & 87.2 & 52.4 & 65.5 \\
\midrule
\multicolumn{7}{@{}l}{\emph{Homogeneous 3$\times$ qwen3.6-27b ($n{=}200$)}}\\
\textbf{Deliberation}    & \textbf{75.6} & 72.5 & \textbf{78.1} & \textbf{82.9} & \textbf{60.3} & 69.8 \\
\quad Union            & 75.4 & \textbf{73.5} & 77.4 & 78.5 & 63.3 & \textbf{70.1} \\
\quad Majority vote    & 67.3 & 54.5 & 72.0 & 88.9 & 37.9 & 53.2 \\
\bottomrule
\end{tabular}
\caption{Moderated consensus vs.\ union and majority-vote aggregation of
independent Phase 1 juror judgements, for the two homogeneous (3 identical
jurors, same-model moderator) panels in the \emph{Diversity of the panel
matters} strip figure (\Cref{sec:results:diversity} main text). For gpt-oss-120b,
deliberation improves on union across every metric (largest gain: recall
0.741~$\to$~0.786). For qwen3.6-27b --- the weakest juror model in this
report --- union's Bal.~F1 (0.735) and F1 (0.701) both \emph{exceed}
consensus's (0.725, 0.698): deliberation loses more true positives than it
gains in precision for this panel, the only reversal of the consensus-beats-union
pattern we observe across all six panels in this appendix. Consensus's Bal.~Acc
(0.756) is likewise essentially tied with union's (0.754).}
\label{tab:delib-vs-agg-diversity}
\end{table}

Deliberation's effect on Bal-F1 itself is systematically
different depending on what the panel needs: for the frontier panel, where gpt-5.4 alone nearly saturates the task, unioning in four additional jurors' raw claims increases
recall above consensus (88.4 vs. 87.1) but at a precision cost (72.2 vs. 77.5), so deliberation trades a small amount of that recall back for a larger precision
gain.
  
For panels dominated by weaker jurors (Large OSS, Small/Med OSS), the same effect appears in milder form, where recall is slightly degraded and precision is higher in deliberation compared to union (e.g. Small/Med OSS: +4.1 precision, $-$1.1 recall, net +1.4 Bal-F1); and for a single strong model resampled three times (3$\times$
gpt-oss-120b), deliberation instead acts as a recall-recovery mechanism, surfacing defects that no individual sample caught while precision is unchanged (74.1$\rightarrow$78.6 recall at
constant $\sim$81\% precision). This adaptivity (pruning spurious claims when the panel is noisy, trading recall for precision when a dominant juror over-inflates the union, and
recovering missed claims when the panel is merely under-sampled, all on top of a consolidated and evidence-backed final verdict) is a capability neither union nor majority vote rule can offer.  Deliberation instead diagnoses and corrects the specific
failure mode present in a given panel's raw judgements, at the cost of the additional inference required to reach that single verdict. These behaviors may be partly an artifact of Hard2Verify itself: on a benchmark where jurors are already reasonably well-calibrated defect detectors,
the marginal false positive is more common than the marginal false negative, so pruning has more room to help than recovery does. On a harder task where models struggle to identify defects at all we would expect this balance to flip, with deliberation's recall-recovery role (as seen here only in the homogeneous gpt-oss-120b panel) becoming the dominant source of benefit rather than the exception.

\section{On the difficulty of Hard2Verify}
\label{app:h2vdifficulty}

We chose Hard2Verify \citep{lin2025hard2verify} as our benchmark because it is, to our
knowledge, the hardest and most contamination-resistant publicly available step-level
reasoning-defect dataset. Two properties support this: the \emph{source} of its problems (which
governs both intrinsic difficulty and the chance of train--test contamination) and the
\emph{length} of the reasoning being verified (a segmentation-independent proxy for how much
there is to get wrong). We contrast it throughout with ProcessBench \citep{zheng2024processbench},
the closest prior step-level error-detection benchmark.

\paragraph{Data source: difficulty and contamination.}
ProcessBench draws its problems from four established public test sets: GSM8K
\citep{cobbe2021gsm8k} (grade-school arithmetic), MATH (competition), and OlympiadBench and
Omni-MATH (olympiad). Two of these tiers are effectively saturated by modern models, and---because
the problems have been public for years and are widely used for training---even the harder tiers
carry a real risk of train--test contamination: a verifier may have seen a given problem (and its
canonical solution) during pre-training, so strong verification scores can partly reflect memorised
solutions rather than genuine step-by-step checking. Hard2Verify is constructed to remove both
issues. Its problems are drawn from \emph{very recent} open-ended competition and olympiad
mathematics---the regime in which frontier LLM systems only reached gold-medal level at IMO
2025---chosen to postdate typical training cutoffs, so a high score is much harder to obtain by
recall. The solutions being verified are themselves generated by \emph{frontier} models on
open-ended problems, rather than reformatted outputs of smaller open-source models. The result is a
benchmark whose difficulty is uniformly at the frontier, with no easy floor and a reduced
contamination surface.

\paragraph{Record length.}
Length is a difficulty signal that does not depend on how a solution is segmented into steps: a
longer solution has more places for a subtle error to hide and a longer dependency chain a verifier
must follow. We measured the length of every solution being verified---in tokens, using a
byte-level BPE tokenizer---for Hard2Verify and for each ProcessBench subset; \Cref{tab:h2v-length}
reports the distributions. Two patterns stand out. First, within ProcessBench length tracks
difficulty exactly as expected, rising from $\sim260$ tokens on grade-school GSM8K to $\sim760$
on the olympiad-level subsets. Second, Hard2Verify solutions are far longer than any ProcessBench
tier: a mean of $\sim2{,}000$ tokens (median $\sim1{,}400$), roughly $2.7\times$ the
olympiad-level ProcessBench subsets and nearly $8\times$ grade-school GSM8K, with a heavy tail
(90th percentile $\sim3{,}900$ tokens, maximum over $12{,}000$). ProcessBench's \emph{longest}
olympiad solution is close to Hard2Verify's \emph{median}. Longer, frontier-generated, open-ended
reasoning is precisely the regime a single judge struggles to verify reliably, and where a
deliberating jury has the most to add.

\begin{table}[!htbp]
\centering
\small
\caption{Length of the solution being verified, in byte-level BPE tokens, for Hard2Verify and each
ProcessBench subset. Hard2Verify solutions are markedly longer than even the olympiad-level
ProcessBench subsets, a segmentation-independent indication of its higher difficulty ceiling.
Absolute counts depend on the tokenizer; the cross-dataset \emph{ratios} do not.}
\label{tab:h2v-length}
\begin{tabular}{@{}lrrrrr@{}}
\toprule
\textbf{Dataset} & \textbf{$n$} & \textbf{Mean} & \textbf{Median} & \textbf{P90} & \textbf{Max} \\
\midrule
\textbf{Hard2Verify} & $200$ & $\mathbf{2008}$ & $\mathbf{1383}$ & $3924$ & $12577$ \\
\midrule
ProcessBench / GSM8K (grade-school) & $400$ & $264$ & $248$ & $391$ & $917$ \\
ProcessBench / MATH (competition)   & $1000$ & $510$ & $430$ & $968$ & $1897$ \\
ProcessBench / OlympiadBench (olympiad) & $1000$ & $759$ & $700$ & $1235$ & $1967$ \\
ProcessBench / Omni-MATH (olympiad) & $1000$ & $755$ & $715$ & $1213$ & $2142$ \\
\bottomrule
\end{tabular}
\end{table}

\section{Defect taxonomy and per-subcategory fingerprints}
\label{app:defect-taxonomy}

This appendix gives the methodology behind the shared defect taxonomy used in \Cref{sec:profiling}, the taxonomy itself, and the full per-subcategory fingerprint table.

\paragraph{How the taxonomy is built.}
The taxonomy is \emph{emergent} (induced from the data, not hand-authored) and \emph{shared} (induced once over the pooled defects of all three models, so every model's distribution is over the same categories). It is built by a single LLM in three passes over the pooled defect set. \textbf{(1)~Grow.} The pooled defects are deterministically interleaved so each batch mixes models, then processed in batches. Each grow step shows the LLM the current taxonomy plus a new batch of defect descriptions and asks for the updated taxonomy under a strict rule: preserve every existing category and subcategory verbatim, and add a node only when a defect fits nothing existing. This is accretion rather than rewrite, so the taxonomy grows monotonically and converges as batches stop introducing novel error kinds. \textbf{(2)~Consolidate.} One merge pass dedupes near-identical nodes, keeps the set mutually exclusive and collectively exhaustive, and freezes the two-layer (category $\rightarrow$ subcategory) shape into a flat list of leaves with stable ids. \textbf{(3)~Assign.} A final pass labels every defect against the frozen leaves; assignment first resets any prior label so an omitted defect stays explicitly unassigned rather than retaining a stale id, and leftover unassigned defects are retried until covered. The run reported here yields \textbf{8 categories and 56 subcategory leaves}. Because grow and consolidate are LLM passes at low but non-zero temperature, a re-run can produce slightly different labels and counts; the taxonomy below is the result of \emph{this} run, not a canonical fixed schema.

\paragraph{The taxonomy.}
The eight top-level categories and their subcategory leaves are:
\begin{description}[leftmargin=0pt,itemsep=2pt,topsep=2pt]
  \item[Computation \& algebraic manipulation (8).] Numerical calculation error; misusing a previously computed or given value; modular arithmetic error; other arithmetic/computation error; sign error; incorrect expansion, simplification, or factoring; incorrect formula, identity, or substitution; invalid or mislabeled manipulation.
  \item[Logical reasoning (8).] Incorrect claim about a mathematical fact; faulty logical inference (quantifiers, necessity/sufficiency); failure to handle contradictions or dismiss/keep cases correctly; conflating distinct concepts; incorrect example construction or verification; reversing direction of inequality or bound; incorrect independence or factorization assumptions; other logical-reasoning error.
  \item[Combinatorial/Structural reasoning (6).] Incorrect recurrence or counting formula; double-counting or overcounting; missing cases or regions in decomposition; unjustified structural or completeness claims; invalid base cases or boundary conditions; incomplete enumeration of valid configurations.
  \item[Geometric/Spatial reasoning (7).] Incorrect coordinate or vector computations; flawed geometric arguments or property claims; incorrect transformation application; spatial orientation or ordering error; applying geometric formulas to invalid configurations; misidentifying geometric shapes or regions; other geometric/spatial error.
  \item[Problem interpretation (6).] Misreading or altering given conditions; substituting fabricated conditions for stated ones; incorrect mathematical modeling or constraint formulation; omitting stated constraints from the solution; misidentifying structural or geometric roles; misreading diagrams or code specifications.
  \item[Proof methodology and rigor (8).] Unproven assumptions used as established facts; incomplete case analysis or verification; overgeneralization or ad-hoc argument as proof; answer asserted without derivation (guessing or memorized recall); insufficient or erroneous bounds/estimates; circular or self-referential verification; applying proof techniques with violated prerequisites; approximate numerical evidence treated as exact proof.
  \item[Self-monitoring and error-handling (7).] Extended reasoning on an uncorrected false premise; self-corrected error that does not propagate; failure to diagnose the source of a detected error; multiple contradictory values left unresolved; unsupported final answer disconnected from derivation; final answer contradicting correct intermediate work; other self-monitoring/error-handling failure.
  \item[Output/generation failures (6).] Truncated or incomplete solution; infinite loop or cycling without progress; verbose redundancy without progress; abandoning the problem for alternative interpretations; abandoning a productive approach without resolution; degenerate or garbled output.
\end{description}

\section{nemotron-3-super: subcategory distribution and worked examples}
\label{app:profiling:nemotron-examples}

This appendix supports the close read of nemotron-3-super in \Cref{sec:profiling:fingerprints}: the full subcategory (leaf) breakdown of its defects, and a worked example of a single trace with defects at each severity level.

\begin{figure}[!htbp]
\centering
\includegraphics[width=0.85\linewidth]{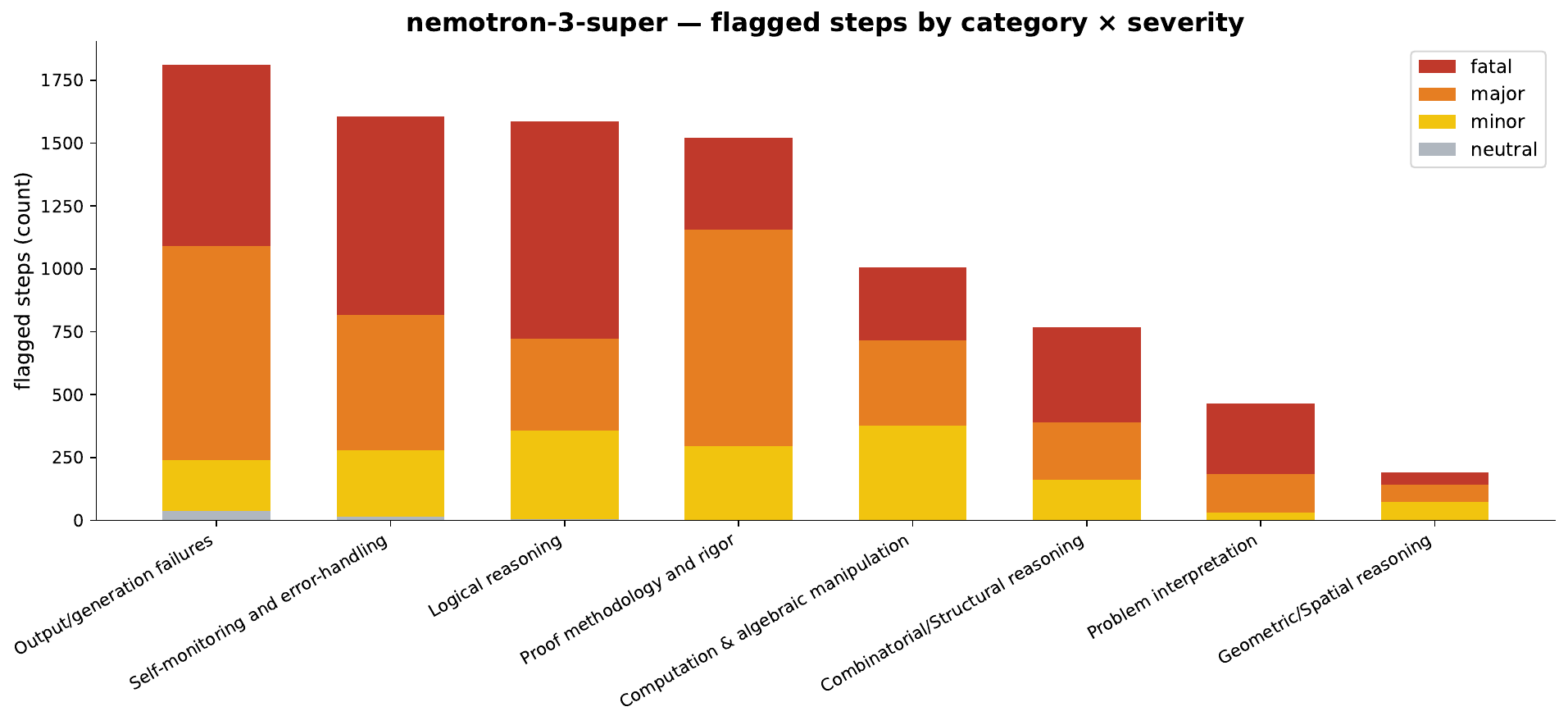}
\caption{nemotron-3-super: flagged reasoning steps by taxonomy category, stacked by severity (regex segmentation, AIME2026, $1{,}920$ traces). Output/generation failures and self-monitoring/error-handling are the two largest categories by flagged-step count; logical reasoning is the most fatal-heavy, whereas proof methodology and rigor leans major/minor (lapses in rigor rather than false claims).}
\label{fig:nemotron-stacked}
\end{figure}

\subsection{Subcategory (leaf) distribution}
\label{app:profiling:nemotron-subcat}

\Cref{tab:nemotron-subcat} lists every non-zero taxonomy leaf for nemotron-3-super, grouped by parent category and sorted by defect count within category. Of the $56$ leaves in the shared taxonomy, $54$ are populated by at least one of nemotron-3-super's $3{,}384$ defects (a defect may cite several steps; this table counts defects, not flagged steps, so totals differ from \Cref{fig:nemotron-stacked}, which counts flagged steps). One leaf dominates: \emph{extended reasoning on an uncorrected false premise} ($927$ defects, $27.4\%$ of the model's defects), accounting for the bulk of the fatal mass seen in the self-monitoring category of \Cref{fig:nemotron-stacked}; the next-largest leaves are \emph{truncated or incomplete solution} ($254$, $7.5\%$) and \emph{incorrect claim about a mathematical fact} ($218$, $6.4\%$).

{\small
\setlength{\tabcolsep}{5pt}
\begin{longtable}{@{}p{0.62\linewidth}rr@{}}
\caption{nemotron-3-super: every populated taxonomy leaf, grouped by category and sorted by defect count within category. Percentages are of the model's $3{,}384$ total defects. Regex segmentation, AIME2026, $1{,}920$ traces.}
\label{tab:nemotron-subcat}\\
\toprule
Subcategory & Count & \% of defects \\
\midrule
\endfirsthead
\toprule
Subcategory & Count & \% of defects \\
\midrule
\endhead
\midrule
\multicolumn{3}{r}{\emph{continued on next page}}\\
\endfoot
\bottomrule
\endlastfoot
\multicolumn{3}{@{}l}{\textit{Computation \& algebraic manipulation (476 defects)}}\\
Numerical calculation error & $158$ & $4.7\%$ \\
Incorrect formula, identity, or substitution & $103$ & $3.0\%$ \\
Misusing a previously computed or given value & $99$ & $2.9\%$ \\
Invalid or mislabeled manipulation & $48$ & $1.4\%$ \\
Sign error & $27$ & $0.8\%$ \\
Incorrect expansion, simplification, or factoring & $26$ & $0.8\%$ \\
Modular arithmetic error & $15$ & $0.4\%$ \\
\midrule
\multicolumn{3}{@{}l}{\textit{Logical reasoning (554 defects)}}\\
Incorrect claim about a mathematical fact & $218$ & $6.4\%$ \\
Conflating distinct concepts & $104$ & $3.1\%$ \\
Failure to handle contradictions or dismiss/keep cases correctly & $82$ & $2.4\%$ \\
Faulty logical inference --- quantifiers, necessity/sufficiency & $61$ & $1.8\%$ \\
Incorrect independence or factorization assumptions & $36$ & $1.1\%$ \\
Other logical-reasoning error & $27$ & $0.8\%$ \\
Incorrect example construction or verification & $17$ & $0.5\%$ \\
Reversing direction of inequality or bound & $9$ & $0.3\%$ \\
\midrule
\multicolumn{3}{@{}l}{\textit{Combinatorial/Structural reasoning (228 defects)}}\\
Unjustified structural or completeness claims & $60$ & $1.8\%$ \\
Incorrect recurrence or counting formula & $54$ & $1.6\%$ \\
Missing cases or regions in decomposition & $48$ & $1.4\%$ \\
Double-counting or overcounting & $45$ & $1.3\%$ \\
Incomplete enumeration of valid configurations & $13$ & $0.4\%$ \\
Invalid base cases or boundary conditions & $8$ & $0.2\%$ \\
\midrule
\multicolumn{3}{@{}l}{\textit{Geometric/Spatial reasoning (93 defects)}}\\
Flawed geometric arguments or property claims & $44$ & $1.3\%$ \\
Spatial orientation or ordering error & $32$ & $0.9\%$ \\
Incorrect coordinate or vector computations & $8$ & $0.2\%$ \\
Applying geometric formulas to invalid configurations & $4$ & $0.1\%$ \\
Incorrect transformation application & $3$ & $0.1\%$ \\
Misidentifying geometric shapes or regions & $2$ & $0.1\%$ \\
\midrule
\multicolumn{3}{@{}l}{\textit{Problem interpretation (113 defects)}}\\
Misreading or altering given conditions & $36$ & $1.1\%$ \\
Incorrect mathematical modeling or constraint formulation & $31$ & $0.9\%$ \\
Omitting stated constraints from the solution & $16$ & $0.5\%$ \\
Substituting fabricated conditions for stated ones & $14$ & $0.4\%$ \\
Misreading diagrams or code specifications & $14$ & $0.4\%$ \\
Misidentifying structural or geometric roles & $2$ & $0.1\%$ \\
\midrule
\multicolumn{3}{@{}l}{\textit{Proof methodology and rigor (406 defects)}}\\
Incomplete case analysis or verification & $130$ & $3.8\%$ \\
Unproven assumptions used as established facts & $122$ & $3.6\%$ \\
Overgeneralization or ad-hoc argument as proof & $51$ & $1.5\%$ \\
Applying proof techniques with violated prerequisites & $33$ & $1.0\%$ \\
Circular or self-referential verification & $25$ & $0.7\%$ \\
Answer asserted without derivation --- guessing or memorized recall & $23$ & $0.7\%$ \\
Insufficient or erroneous bounds/estimates & $20$ & $0.6\%$ \\
Approximate numerical evidence treated as exact proof & $2$ & $0.1\%$ \\
\midrule
\multicolumn{3}{@{}l}{\textit{Self-monitoring and error-handling (1132 defects)}}\\
Extended reasoning on an uncorrected false premise & $927$ & $27.4\%$ \\
Self-corrected error that does not propagate & $111$ & $3.3\%$ \\
Failure to diagnose the source of a detected error & $59$ & $1.7\%$ \\
Multiple contradictory values left unresolved & $16$ & $0.5\%$ \\
Other self-monitoring/error-handling failure & $8$ & $0.2\%$ \\
Final answer contradicting correct intermediate work & $6$ & $0.2\%$ \\
Unsupported final answer disconnected from derivation & $5$ & $0.1\%$ \\
\midrule
\multicolumn{3}{@{}l}{\textit{Output/generation failures (382 defects)}}\\
Truncated or incomplete solution & $254$ & $7.5\%$ \\
Abandoning a productive approach without resolution & $85$ & $2.5\%$ \\
Verbose redundancy without progress & $21$ & $0.6\%$ \\
Degenerate or garbled output & $10$ & $0.3\%$ \\
Abandoning the problem for alternative interpretations & $7$ & $0.2\%$ \\
Infinite loop or cycling without progress & $5$ & $0.1\%$ \\
\end{longtable}
}

The severity-stacked, top-three-leaves-per-category view of this same data is given as \Cref{fig:nemotron-stacked-subcat} in the main text (\Cref{sec:profiling:fingerprints}).

\subsection{Example defects by severity}
\label{app:profiling:nemotron-worked-example}

To make the taxonomy and severity labels concrete, we walk through one nemotron-3-super record (AIME2026 problem 28, sample 4) on which the jury raised defects at all three non-neutral severities. The problem asks for the minimum size of a set $S$ of integers with exactly $4040$ ``cousins'' $T$ (disjoint, same size, and pairable with elements differing by exactly $1$). The trace runs to $210$ regex-segmented steps and is truncated before reaching a final answer (the correct answer is $107$, from the factorization $4040 = 2^3 \times 5 \times 101$ realised by components of sizes $1,1,1,4,100$).

The jury raises three defects on three different steps:

\medskip\noindent\textbf{\code{fatal}: Truncated or incomplete solution (Output/generation failures).} At \code{[STEP-205]}, the trace establishes a correct component-based counting formula (the number of cousins is the product of $(k_i+1)$ over path components) by \code{[STEP-201]}, but never applies it to the target value $4040$: it does not factor $4040$, does not choose the factorization that minimises $|S|$, and does not construct a concrete set. It terminates mid-sentence (``\,edges: 0-(-1), 1-\,'') while enumerating the example $S=\{0,1,3\}$, and the final steps break off while beginning the $k \ge 2$ case --- so the problem is never answered despite the machinery being in place.

\medskip\noindent\textbf{\code{major}: Unproven assumptions used as established facts (Proof methodology and rigor).} At \code{[STEP-49]}, the proof that the matching from $S$ to $T$ is unique considers only $2$-element swaps and concludes ``you cannot swap assignments and keep $T$ the same unless $s_1=s_2$''; it never rules out longer cycles (e.g.\ a $3$-cycle $s_1{\to}t_2,\,s_2{\to}t_3,\,s_3{\to}t_1$) that could produce the same set $T$. The uniqueness conclusion is in fact correct, but the argument as written has a gap.

\medskip\noindent\textbf{\code{minor}: Unproven assumptions used as established facts (Proof methodology and rigor).} At \code{[STEP-21]}, after examining $S=\{0,1,2\}$ the trace states ``$S$ cannot have cousins if it contains consecutive numbers'', which is too strong; $S$ may contain consecutive pairs (e.g.\ $S=\{0,1\}$ has cousin $T=\{-1,2\}$). The correct restriction, which the trace itself states one step later in \code{[STEP-22]}, is that $S$ contains no three consecutive integers; the misstatement is immediately self-corrected.
\medskip

\section{Reproducibility notes}
\label{app:repro}

The framework supports six model providers (AWS Bedrock, OpenAI, Google Gemini,
OpenRouter, OpenAI-compatible local clusters, and a Bedrock Mantle gateway),
selected per model ID by prefix. Per-juror clients use separate connection pools
to avoid TLS-handshake races under parallel fan-out. The validation harness is
parallel, resumable, and logs every LLM call (caller, phase, model, prompts,
response, reasoning text, token usage) for audit.

\section{DeltaBench: full results and per-domain analysis}
\label{app:deltabench}

This appendix expands the DeltaBench generalization result
(\Cref{sec:results:deltabench}) with the complete metric set under both scoring
protocols, the full aggregation ladder, and per-domain breakdowns.

\paragraph{Setup.}
DeltaBench \citep{he-etal-2025-large} provides $1{,}236$ long chain-of-thought
traces from o1-style generators (QwQ, DeepSeek-R1) with human annotations at the
\emph{section} level (a section groups consecutive steps addressing one
sub-task). We map each annotated section to one \code{[STEP-k]} marker of our
input format, discarding DeltaBench's finer step indexing, and run our pipeline
unchanged (standard defect prompt, minor$+$ threshold, defect propagation on).
Ground truth follows DeltaBench's own scoring convention: a section is positive
if it appears in either the annotated \emph{error} list or the annotated
\emph{unuseful} list (their published evaluator unions the two). We report two
metric families side by side:

\begin{itemize}[leftmargin=1.4em,itemsep=2pt,topsep=2pt]
  \item \textbf{Ours}: the corpus-level step metrics used throughout this
  paper: Balanced Accuracy, Balanced F1, Accuracy, Precision, Recall, and F1,
  with error as the positive class, pooled over all steps of all records.
  \item \textbf{Theirs}: DeltaBench's protocol, replicated exactly on our
  structured predictions: per-sample precision/recall/F1 over the set of
  predicted vs.\ annotated error sections, with predictions beyond the last
  annotated section discarded, then macro-averaged across samples
  (F1$_\text{M}$); micro variants pool TP/FP/FN over samples before computing
  the ratios. Because our verdicts already carry structured step citations, no
  LLM-based answer extraction is involved.
\end{itemize}

The jury is the homogeneous $3\times$ gpt-oss-120b panel with a same-model
moderator; the single-model comparator is one opus-4.6 Phase 1 pass with the
identical prompt. All rows are computed on the same $1{,}226$ records
(``paired''): opus could not complete $10$ of the $1{,}236$ traces (the
longest in the benchmark) returning empty responses even at a
$24{,}000$-token output ceiling after multiple retries; those records are
excluded from both systems. Since the excluded traces are concentrated in the
hardest math records, the pairing, if anything, favors the single-model
baseline.

\paragraph{Overall results.}
\Cref{tab:deltabench-overall} reports the full ladder. There are three main observations.
First, the deliberated consensus beats every single system (except gpt-5.4 where the performance is very close) under both metric
families: $+5.8$ Balanced F1 over the best individual juror and $+3.1$ over
opus solo under our metrics; $+8.2$ and $+4.5$ macro-F1 respectively under
theirs. Second, the aggregation ladder replicates the Hard2Verify pattern
exactly: majority vote barely improves on a single juror ($38.7$ macro-F1 vs.\
$39.7$ best solo), the union of Phase 1 judgements captures most of the
available gain ($48.2$), and deliberation lands beside the union ($47.9$) while
achieving the highest recall of any method ($65.7$ macro-recall) --- on this
benchmark, at this severity threshold, deliberation and union are near-equivalent
on F1 and differ in their precision/recall trade-off. Third, for leaderboard
context, the strongest critic reported by \citet{he-etal-2025-large} is
gpt-4-turbo at $40.8$ macro-F1: the jury exceeds it by $7$ points using three
small open-weight models, and even our opus-4.6 single-model baseline ($43.4$)
exceeds it, indicating that part of the advantage comes from our defect-prompt
format and propagation convention, and the rest ($+4.5$) from deliberation.

\begin{table}[h]
\centering
\small
\renewcommand{\arraystretch}{1.1}
\setlength{\tabcolsep}{3.4pt}
\begin{tabular}{@{}l rrrrrr rrrr@{}}
\toprule
& \multicolumn{6}{c}{\textbf{Ours (step-level, corpus)}} & \multicolumn{4}{c}{\textbf{DeltaBench protocol}} \\
\cmidrule(lr){2-7} \cmidrule(l){8-11}
\textbf{Method} & Bal.~Acc & Bal.~F1 & Acc. & Prec. & Rec. & F1 & F1$_\text{M}$ & P$_\text{M}$ & R$_\text{M}$ & F1$_\mu$ \\
\midrule
gpt-oss seat 1 & 59.4 & 53.3 & 74.1 & 19.5 & 40.4 & 26.3 & 37.9 & 39.8 & 43.2 & 40.1 \\
gpt-oss seat 2 & 60.5 & 55.2 & 74.3 & 20.2 & 42.6 & 27.4 & 39.7 & 40.5 & 46.4 & 41.0 \\
gpt-oss seat 3 & 60.5 & 55.7 & 73.6 & 19.9 & 43.5 & 27.3 & 39.7 & 40.0 & 46.9 & 41.7 \\
Majority(3) & 60.8 & 54.0 & 76.4 & 21.6 & 40.5 & 28.2 & 38.7 & 40.3 & 44.5 & 41.3 \\
Union(3) & 61.7 & 61.7 & 61.9 & 17.2 & 61.5 & 26.9 & \textbf{48.2} & 44.8 & 64.1 & 45.4 \\
\textbf{Jury deliberation} & 61.6 & 61.5 & 60.0 & 16.8 & \textbf{63.6} & 26.6 & 47.9 & 43.7 & \textbf{65.7} & 45.2 \\
opus-4.6 solo & 59.8 & 58.4 & 66.7 & 17.3 & 50.7 & 25.8 & 43.4 & 42.8 & 54.1 & 44.9 \\
gpt-5.4 solo & 61.8 & \textbf{61.8} & 61.3 & 17.1 & 62.5 & 26.9 & 47.2 & 42.9 & 65.6 & \textbf{46.2} \\
\bottomrule
\end{tabular}
\caption{DeltaBench, overall ($n{=}1{,}226$ paired records). Left block: our corpus-level step metrics. Right block:
DeltaBench's own protocol (per-sample macro-averaged F1/precision/recall and
micro-F1). The deliberated consensus beats every individual system except
gpt-5.4 solo, which lands beside it under both metric families ($61.8$ vs.\
$61.5$ Balanced F1; $47.2$ vs.\ $47.9$ F1$_\text{M}$) with an essentially
identical recall profile --- replicating the Hard2Verify finding that gpt-5.4
alone saturates what the panel achieves. Union and consensus are
near-equivalent on F1, with consensus taking the highest recall. DeltaBench's
best published critic scores $40.8$ F1$_\text{M}$ \citep{he-etal-2025-large}.}
\label{tab:deltabench-overall}
\end{table}

\section{Defect-guided retry with jury feedback}
\label{app:retry}

This section details the downstream evaluation summarized in
\Cref{sec:profiling:retry}. Whereas the profiling analysis characterizes the
types of defects produced by a reasoning model, this experiment asks whether
the jury's structured findings help that model correct its own solutions. We
evaluate retries against the AIME answer key, which provides an external
success criterion independent of the jury.

\subsection{Design}
\label{app:retry:design}

\paragraph{Data.}
For each of the $30$ AIME2026 problems, we generate $64$ fresh solutions with
nemotron-3-super, for a total of $1{,}920$ traces. Generation uses temperature
$0.7$, a maximum of $32{,}000$ tokens, and high reasoning effort, matching
\Cref{sec:profiling:setup}. All generation calls complete successfully, and
exact-answer scoring gives a corpus pass@1 of $80.0\%$
($1{,}536/1{,}920$).

We segment every trace with the regular-expression procedure of
\Cref{sec:method:steps} and evaluate it with the profiling jury: glm-5.2,
minimax-m3, deepseek-v4-pro, and kimi-k2.6, with deepseek-v4-pro as moderator.
The jury produces a deliberated verdict for every trace and identifies
$3{,}366$ non-neutral defects. In total, $877$ traces ($45.7\%$) contain at
least one defect with minor, major, or fatal severity. We exclude neutral
findings from the retry prompts because they do not identify an actionable
error. Among the $384$ wrong-answer traces, the resulting gate flags $372$
($96.9\%$ recall); the remaining $12$ receive no non-neutral finding.

\paragraph{Retry conditions.}
The same nemotron-3-super deployment retries every flagged trace at temperature
$0.7$, with a maximum of $32{,}000$ generation tokens and high reasoning
effort. Each trace is evaluated under two conditions:

\begin{description}[leftmargin=1.2em,itemsep=3pt,topsep=2pt]
  \item[\code{full} (rich feedback).] The model receives the problem, its
    previous \code{[STEP-k]}-marked trace, and every non-neutral jury finding.
    Each finding includes its cited step locations, severity, and verbatim
    \code{what\_went\_wrong} diagnosis.
  \item[\code{steps} (anchors only).] The model receives the same problem and
    previous trace, but the findings are reduced to a deduplicated list of
    cited \code{[STEP-k]} locations, with no severity labels or explanations.
\end{description}

This design yields paired outcomes for all $877$ flagged traces and $1{,}754$
retry generations in total. Every retry call completes, although $49$
\code{steps} outputs and $27$ \code{full} outputs contain no parseable integer
and are scored as incorrect. We first extract an answer from an explicit
\code{ANSWER} marker and only then fall back to the reasoning text; this
prevents a number in a truncated reasoning tail from being accepted as the
answer. Both prompts instruct the model to re-solve the problem from scratch
rather than patch individual steps. The paired contrast therefore measures
the incremental value of diagnosis and severity when both conditions already
provide the same localization signal.

\subsection{Prompts}
\label{app:retry:prompts}

The conditions use the same prompt scaffold and differ only in the
review-findings block. The \code{steps} prompt is reproduced below.
Placeholders in braces are filled separately for each trace; when a finding
cites multiple steps, every cited location is included in the deduplicated
list.

\begin{lstlisting}[caption={The \code{steps} retry prompt.},captionpos=b]
You previously attempted the following competition problem.

## Problem
{problem}

## Your previous solution
{marked_trace}

## Review findings
An automated review of your solution flagged issues at the
following steps:

- [STEP-14]
- [STEP-15]

## Task
Re-solve the problem from scratch. Use the flagged locations to
decide which parts of your previous approach to distrust; do not
assume unflagged steps are correct. Show your reasoning, then end
with:
ANSWER: <integer>
\end{lstlisting}

The \code{full} prompt replaces the location list with complete findings that
include severity and the jury's explanation:

\begin{lstlisting}[caption={The \code{full} arm's findings block (one entry shown).},captionpos=b]
- [STEP-14] -- FATAL
  "In [STEP-14], the trace writes an incorrect intermediate
   expression: 'right: 18 * (2v(v+9)/9) = 4 * 2 v(v+9)' ..."
\end{lstlisting}

Because every AIME answer is an integer in $[0,999]$, the
\code{ANSWER: <integer>} contract matches the benchmark's answer format.
Jury explanations sometimes state an explicit correction or corrected value.
Accordingly, \code{full} represents the complete actionable feedback produced
by the jury, including such corrections when present; \code{steps} isolates
the information supplied by localization alone.

\subsection{Results}
\label{app:retry:results}

\paragraph{Outcome transitions.}
\Cref{tab:retry-flips} separates beneficial transitions (an initially wrong
answer becomes correct) from regressions (an initially correct answer becomes
wrong). Rich feedback improves both sides of this tradeoff: it corrects
$48.7\%$ of the $372$ flagged wrong answers, compared with $39.2\%$ for step
anchors, while reducing the regression rate from $5.3\%$ to $3.6\%$ among the
$505$ flagged correct answers.

\begin{table}[h]
\centering
\small
\begin{tabular}{@{}lcccc@{}}
\toprule
Condition & Traces & Fixes (wrong$\to$right) & Regressions (right$\to$wrong) & Net \\
\midrule
\code{steps} & $877$ & $146/372$ ($39.2\%$) & $27/505$ ($5.3\%$)  & $+119$ \\
\code{full}  & $877$ & $181/372$ ($48.7\%$) & $18/505$ ($3.6\%$)  & $+163$ \\
\bottomrule
\end{tabular}
\caption{Retry outcome transitions by feedback condition ($64$ samples per
problem). The non-neutral-defect gate selects $372$ of the $384$ wrong-answer
traces, so both conditions cover $96.9\%$ of wrong answers while retrying only
$877$ of the $1{,}920$ traces. Net is the number of fixes minus regressions.}
\label{tab:retry-flips}
\end{table}

\paragraph{Matched comparison.}
Before retry, accuracy on the $877$ flagged traces is $57.6\%$
($505/877$). Step anchors raise observed retry accuracy to $71.2\%$
($624/877$), whereas rich feedback raises it to $76.2\%$ ($668/877$).
Relative to anchors alone, rich feedback therefore adds $5.0$ percentage
points, comprising $35$ additional fixes and nine fewer regressions.

The paired outcomes are discordant on $112$ traces: rich feedback alone is
correct on $78$, whereas step anchors alone are correct on $34$. An exact
McNemar test gives $p=4\times10^{-5}$. Because the $64$ traces drawn for each
problem are not independent, we also compute paired differences at the
problem level. The mean improvement is $+4.1$ percentage points, with a
bootstrap $95\%$ confidence interval of $[+2.3,+6.1]$. Among the $29$
problems with at least one flagged trace, \code{full} outperforms \code{steps}
on $15$, underperforms on one, and ties on $13$. Both analyses support a
positive incremental effect from the richer feedback package. They do not,
however, separate the contribution of \code{what\_went\_wrong} from that of
severity.

\paragraph{Defect-gated retry policies.}
The experiment also supports a simple selective-retry policy: retry a trace
when the jury reports a qualifying defect, and otherwise retain the original
answer. The table below reports corpus accuracy over all $1{,}920$ traces,
whose original pass@1 is $0.800$.

\begin{center}
\small
\begin{tabular}{@{}lcc@{}}
\toprule
Policy & Corpus accuracy & Retries \\
\midrule
Any non-neutral defect $\to$ \code{full} & $\mathbf{0.885}$ & $877$ \\
Any non-neutral defect $\to$ \code{steps} & $0.862$ & $877$ \\
Fatal defect only $\to$ \code{full} & $0.853$ & $338$ \\
Fatal defect only $\to$ \code{steps} & $0.835$ & $338$ \\
\bottomrule
\end{tabular}
\captionof{table}{Corpus accuracy under selective retry policies. Traces that
do not satisfy the gate retain their original answers.}
\label{tab:retry-policies}
\end{center}

With the any-defect gate, the system retries $45.7\%$ of the corpus while
retaining the original answers for the $1{,}043$ unflagged traces. Combining
this gate with rich feedback yields $88.5\%$ corpus accuracy, $8.5$ percentage
points above the original pass@1. This absolute improvement includes any
generic benefit from a second attempt because the experiment has no blind
retry condition. The comparison under the shared gate is more direct: rich
feedback exceeds step anchors by $2.3$ percentage points on the full corpus
($0.885$ vs.\ $0.862$).

Restricting retries to fatal findings reduces the number of retries from $877$
to $338$, or $38.5\%$ of the any-defect volume. With rich feedback, this policy
still reaches $85.3\%$ accuracy, $5.3$ percentage points above pass@1 but
$3.2$ points below the any-defect policy. The difference reflects
wrong-answer traces whose only findings have minor or major severity and are
therefore not retried by the fatal-only gate.

\subsection{Limitations}
\label{app:retry:caveats}

\begin{enumerate}[leftmargin=1.6em,itemsep=3pt,topsep=3pt]
  \item The experiment covers one generator, one benchmark, and one retry
    sample per condition and trace. The per-problem analysis addresses
    dependence among traces from the same problem, but broader
    generalization remains untested.
  \item There is no blind-retry condition. Some improvement over the original
    answers may therefore come from making a second attempt rather than from
    the feedback itself. The paired \code{full}-versus-\code{steps}
    comparison controls for this shared retry opportunity.
  \item The \code{full} condition bundles diagnosis and severity, and some
    diagnoses contain explicit corrections. The experiment consequently
    measures the complete rich-feedback package; it does not isolate the
    effect of either field or distinguish explanation from a supplied fix.
  \item The gate's $96.9\%$ wrong-answer recall is specific to this corpus and
    does not show that each flagged defect caused the corresponding wrong
    answer. Failures caused by omissions may also escape detection.
  \item The jury findings are not human-validated ground truth. The external
    AIME answer key validates the correctness of retry outcomes, not the
    factual accuracy or severity of individual defect descriptions.
\end{enumerate}

\subsection{Artifacts}
\label{app:retry:artifacts}

The traces, per-trace deliberated verdicts, and extracted defects are stored in
\path{cdd_artifacts/defect_profiling/aime2026_nemotron64_regex/}. Retry
generations are stored as one JSON file per trace and condition, together with
the extracted answer, correctness label, and prompt metadata, under
\path{retries/\{steps,full\}/}. The pipeline wrapper
\code{run\_nemotron64.py}, retry harness \code{retry64.py}, and run logs
\code{run.log} and \code{retry64.log} are located in
\path{cdd_artifacts/defect_profiling/}.

\end{document}